\documentclass{article}

\PassOptionsToPackage{numbers, sort, compress}{natbib}
\usepackage[preprint]{neurips_2026}

\usepackage[utf8]{inputenc} 
\usepackage[T1]{fontenc}    
\usepackage{hyperref}       
\usepackage{url}            
\usepackage{booktabs}       
\usepackage{amsfonts}       
\usepackage{nicefrac}       
\usepackage{microtype}      
\usepackage{xcolor}         
\usepackage{amsthm}       
\usepackage{amssymb} 
\usepackage{graphicx}
\usepackage{subcaption}
\usepackage{mathtools}
\usepackage{multirow}

\graphicspath{{figures/}}

\newtheorem{lemma}{Lemma}
\newtheorem{theorem}{Theorem}
\newtheorem{proposition}{Proposition}

\newcommand{\prx}[1]{\mathbb{P}{\left[#1\right]}}

\newcommand{\mc}[1]{\mathcal{#1}}
\newcommand{\mr}[1]{\mathrm{#1}}
\newcommand{\ms}[1]{\mathsf{#1}}
\newcommand{\mb}[1]{\mathbf{#1}}

\title{X-CoSD: Communication-Efficient Cross-Vocabulary Collaborative Speculative Decoding}

\author{%
	Jaeduk~Lee\\
	Seoul National University\\
	Dept. of Electrical and Computer Engineering\\
	\texttt{ljds1224@snu.ac.kr} \\
	\And
	Wan~Choi\\
	Seoul National University\\
	Dept. of Electrical and Computer Engineering\\
	\texttt{wanchoi@snu.ac.kr} \\
}

\begin{document}

\maketitle

\begin{abstract}
	This paper investigates collaborative speculative decoding (CoSD), a distributed large language model (LLM) inference framework in which  an on-device small language model (SLM) drafts candidate tokens and a server LLM verifies them.
	Existing CoSD methods assume a shared vocabulary between the SLM and the LLM and incur substantial communication load because residual resampling requires token distribution exchange between the user device and the edge server.
	To address these limitations, we propose cross-vocabulary CoSD (X-CoSD), a lossless and communication-efficient CoSD framework for heterogeneous SLM--LLM vocabularies.
	X-CoSD is built on hybrid resampling (HR), which splits residual resampling across the common-vocabulary region on the device and the LLM-only region on the server, so that distribution transmission is required only for the common-vocabulary region.
	We further propose X-CoSD-E, an enhanced variant based on server resampling with device verification (SR-DV), in which the server sends only replacement candidates sampled from the server LLM and their corresponding probabilities for local verification at the device.
	We prove that both X-CoSD and X-CoSD-E preserve the server LLM distribution, and experiments show that they significantly improve token generation speed while maintaining generation quality comparable to that of the server LLM.
\end{abstract}

\section{Introduction}
\label{sec:intro}
Transformer~\cite{vaswani17}-based large language models (LLMs) have demonstrated remarkable success across a wide range of real-world applications, such as code generation, chatbots, and content creation.
Despite this success, their inference latency remains a fundamental bottleneck due to autoregressive decoding (AD), which generates tokens sequentially.
To mitigate this limitation, speculative decoding (SD) has emerged as an effective acceleration framework~\cite{leviathan23, chen23, miao24}.
In SD, a lightweight small language model (SLM) first drafts candidate tokens via AD, and a larger LLM verifies them in parallel.
By offloading the sequential drafting process to the SLM, SD can substantially accelerate decoding while preserving the target LLM distribution. 

Building on SD, \emph{Collaborative SD} (CoSD) has attracted growing attention as a framework for distributed LLM inference, in which an on-device SLM and a server LLM jointly perform SD~\cite{qualcomm23, hao24, oh24, oh25, ning25}.
In CoSD, the user device runs the lightweight drafting step with the on-device SLM, while the edge server performs the more expensive verification step with the server LLM.
CoSD offers several appealing benefits. 
First, it enables lossless decoding with respect to the server LLM, thereby overcoming the performance limitations of the lightweight on-device SLM.
Second, it reduces the number of LLM forward passes at the edge server, lowering server-side inference cost.

However, unlike standard SD on a single device, CoSD is constrained not only by computation but also by \emph{communication} between the device and the server. 
Importantly, the communication overhead arises not merely from transmitting a few token IDs, but from the repeated exchange of probability distributions over the vocabulary.
In particular, when a candidate token is rejected, the resampling step 
in SD requires the SLM and LLM token distributions at the rejected position to be co-located, which makes their exchange unavoidable in the collaborative setting.
Over wireless networks, such high-dimensional messages can dominate end-to-end latency, so communication becomes a practical bottleneck in CoSD~\cite{oh24, oh25, ning25}.

In this context, prior works~\cite{oh24, oh25, ning25} have investigated communication-efficient CoSD frameworks.
For example, Uncertainty-aware Hybrid Language Model (U-HLM)~\cite{oh24, oh25} reduces communication load by verifying only uncertain candidate tokens and, in~\cite{oh25}, further compresses the transmitted SLM distribution using top-$k$ truncation.
However, these approaches might degrade response quality due to skipped verification, and they still rely on uploading the SLM distributions at the uncertain positions to the server.
This device-to-server communication (uplink) is typically a bottleneck in wireless networks due to the power constraints of user devices and limited uplink communication bandwidth~\cite{rochman25}. To address this challenge, the authors in~\cite{ning25} proposed a downlink-based design in which the server sends the LLM distribution only upon rejection, thereby enabling on-device resampling.
This design reduces communication latency by limiting distribution transmission to rejection events and leveraging the typically higher downlink transmission rate in wireless networks.

Despite these advances, prior CoSD works have relied on the critical assumption that the on-device SLM and the server LLM share the same vocabulary.
In practice, this shared-vocabulary assumption is overly restrictive, as on-device SLMs may come from different model families or vendors and therefore use tokenizers that do not match that of the server LLM.
When the SLM and the LLM operate on different vocabularies, 
their token distributions are defined over incompatible token spaces, so the verification and resampling procedures in SD can no longer be applied as-is.
Consequently, existing CoSD frameworks cannot handle heterogeneous-vocabulary settings without modification.

A straightforward extension is to adapt SD methods designed for heterogeneous vocabularies~\cite{timor25, ramakrishnan25} to the collaborative setting.
These methods address vocabulary mismatch in different ways.
In~\cite{timor25}, the authors proposed Token-Level Intersection (TLI), which restricts drafting to the common vocabulary shared by both models and enables lossless decoding without additional training.
In contrast, OmniDraft~\cite{ramakrishnan25} constructs an n-gram cache of cross-vocabulary token mappings and continuously aligns the SLM with the LLM via online knowledge distillation.
However, neither approach is well-suited to CoSD for the following reasons. 
First, OmniDraft requires additional adaptation for each SLM--LLM pair, which is impractical in collaboration settings with diverse on-device SLMs. 
Second, while TLI is training-free and lossless, naively integrating it into CoSD requires the server to transmit the full server LLM distribution upon rejection, resulting in substantial communication load.
In other words, existing CoSD methods assume a shared vocabulary, whereas heterogeneous-vocabulary SD methods do not account for communication overhead in the collaborative setting.
Therefore, a practical CoSD framework must satisfy three key requirements: supporting heterogeneous vocabularies, preserving lossless decoding, and maintaining low communication overhead.

To this end, we propose \emph{Cross-Vocabulary CoSD} (X-CoSD), a communication-efficient CoSD framework for heterogeneous SLM--LLM vocabularies.
Building on a TLI-based formulation, in which on-device drafting is restricted to the common vocabulary shared by the SLM and the LLM, X-CoSD redesigns the rejection-handling step to enable exact resampling without transmitting the full server LLM distribution.
Specifically, X-CoSD is built upon \emph{Hybrid Resampling} (HR), which exploits the structure of the residual distribution induced by TLI to enable distributed resampling.
Upon token rejection at the server, HR decomposes the residual distribution into the common-vocabulary and LLM-only regions, and randomly selects the resampling region according to the relative residual probability mass assigned to each region. If the common-vocabulary region is selected, the user device locally resamples the replacement token from the common-vocabulary region; otherwise, the server resamples it from the LLM-only region.
As a result, the server needs to transmit only the common-region portion of the server LLM distribution, rather than the full distribution, thereby significantly reducing communication overhead.
We further propose X-CoSD-E, a more communication-efficient variant of X-CoSD enabled by \emph{Server Resampling with Device Verification} (SR-DV), in which the server sends only a small number of replacement tokens sampled from the server LLM and their corresponding probabilities, while the user device locally verifies them.
When all replacement candidates are rejected, the method falls back to HR to preserve exact residual resampling.
Overall, X-CoSD and X-CoSD-E enable lossless CoSD under heterogeneous vocabularies while substantially reducing communication overhead.

\section{Preliminaries}\label{sec:collaborative}
\subsection{Collaborative speculative decoding with shared vocabularies}\label{subsec:cosd}
Consider a CoSD framework where an on-device SLM and a server LLM operate on a shared vocabulary $\mc{V}$ and jointly generate responses.
The goal of CoSD is to accelerate generation by allowing the on-device SLM to propose candidate tokens, which are subsequently verified by the server-side LLM. 
Given an input prompt $\mb{d}_0$, the response is generated through iterative verification rounds, each consisting of three steps: \emph{1) drafting on the user device, 2) verification on the edge server, and 3) resampling the rejected token on the user device}.

Let $\mb{d}_t$ denote the current prefix at verification round $t$, which is the concatenation of $\mb{d}_0$ and the tokens generated up to round $t-1$. 
Given $\mb{d}_t$, the user device autoregressively generates $N$ candidate tokens $\mb{x}_t \triangleq [x_{t,1}, \dots, x_{t,N}]$.
Here, $x_{t,n}$ is sampled from the token distribution $p_{t,n}(x|\mb{d}_t, \mb{x}_{t,<n})$ of the on-device SLM, where $\mb{x}_{t,<n} \triangleq [x_{t,1}, \dots, x_{t,n-1}]$. 
The user device then requests server-side verification by transmitting $\mb{x}_t$ and $\{p_{t,n}(x_{t,n}|\mb{d}_t, \mb{x}_{t,<n})\}_{n=1}^N$.
It is worth noting that the user device does not upload the full SLM distribution, but only the generation probabilities of the candidates, which are sufficient for verification under the standard SD acceptance rule.
After receiving the verification request, the server runs the LLM on the concatenated input sequence $[\mb{d}_t, \mb{x}_t]$, and obtains the token distributions $\{q_{t,n}(x|\mb{d}_t,\mb{x}_{t,<n})\}_{n=1}^{N+1}$ in parallel. 
For notational simplicity, we write $p_{t,n}(x)$ and $q_{t,n}(x)$ in place of $p_{t,n}(x|\mb{d}_t, \mb{x}_{t,<n})$ and $q_{t,n}(x|\mb{d}_t, \mb{x}_{t,<n})$, respectively, in the remainder of the paper.
Following conventional SD \cite{chen23,leviathan23}, the server accepts each token $x_{t,n}$ with probability
\begin{align}
	\min\left(1, \frac{q_{t,n}(x_{t,n})}{p_{t,n}(x_{t,n})}\right).
	\label{eq:acceptance}
\end{align}

Depending on the verification result, the subsequent step proceeds as follows.
\begin{itemize}
	\item \textbf{Case 1 (Rejection):} If some candidate tokens are rejected and the first rejected token is $x_{t,n}$, the edge server sends a feedback message, which consists of an accepted length $n_\mr{acc} \triangleq n-1$ and the full LLM distribution $q_{t,n}(x)$ at rejected position $n$, to the user device.
	Given the feedback message from the server, the device replaces the rejected token $x_{t,n}$ with a new token $\tilde{x}_{t,n}$ sampled from the residual distribution $p_\mr{res}(x)$, which is defined as
	\begin{align}
		p_\mr{res}(x) \triangleq \frac{\max(q_{t,n}(x)-p_{t,n}(x),0)}{\sum_{x' \in \mc{V}} \max(q_{t,n}(x')-p_{t,n}(x'),0)},
		\quad \forall x \in \mc{V}.
		\label{eq:p_res}
	\end{align}
	\item \textbf{Case 2 (Full Acceptance):} If all $N$ candidate tokens are accepted, the server samples a bonus token $x_{t, N+1}$ from $q_{t,N+1}(x)$ and sends $x_{t,N+1}$ to the device.
\end{itemize}

CoSD inherits the verification and resampling mechanism of conventional SD, thereby preserving the server LLM distribution, as proved in~\cite{leviathan23,chen23}. 
In addition, CoSD improves communication efficiency via an asymmetric communication design: the user device transmits only candidate tokens and their corresponding probabilities over the uplink, which is typically the bottleneck in wireless networks, while the server LLM distribution is transmitted over the downlink only upon rejection.

\subsection{Naive extension to heterogeneous SLM--LLM vocabularies}
\label{sec:naive}
\begin{figure}[!t]
	\centering
	\includegraphics[width=1.0\columnwidth]{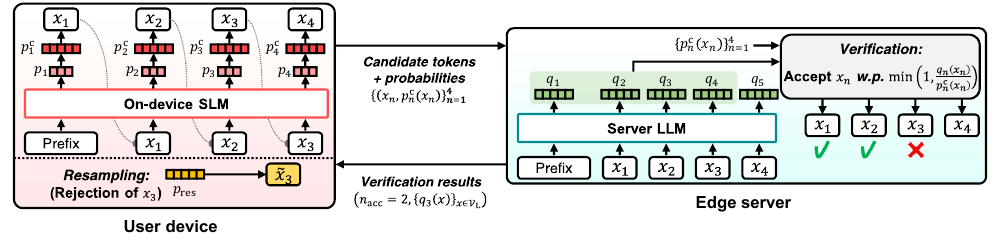}
	\caption{Illustration of the naive extension of CoSD with heterogeneous vocabularies.}
	\label{fig:cosd}
\end{figure}
In CoSD with shared vocabularies, the acceptance criterion and the residual distribution in~\eqref{eq:acceptance} and~\eqref{eq:p_res} rely on $p_{t,n}(x)$ and $q_{t,n}(x)$ being defined over the same vocabulary. 
However, when the SLM and the LLM operate on different vocabularies, their token distributions are not directly comparable, rendering the conventional CoSD framework inapplicable.
Since such cross-vocabulary settings naturally arise in practical device-server collaborative inference, it is essential to develop a CoSD framework that can effectively accommodate heterogeneous vocabularies.

To extend CoSD to heterogeneous SLM--LLM vocabularies in a naive way,  the TLI method proposed in~\cite{timor25} can be utilized.
TLI restricts the candidate token generation to the common vocabulary shared by the SLM and the LLM and guarantees lossless decoding without additional training.
Let $\mc{V}_{\mr{S}}$ and $\mc{V}_{\mr{L}}$ denote the vocabularies of the on-device SLM and the server LLM, respectively. 
The common vocabulary set and the LLM-only vocabulary set are defined as $\mc{V}_{\mr{C}} \triangleq \mc{V}_{\mr{S}} \cap \mc{V}_{\mr{L}}$ and  $\mc{V}_{\mr{O}} \triangleq \mc{V}_{\mr{L}} \setminus \mc{V}_{\mr{C}}$, respectively.
The user device and the edge server are assumed to share the tokenizer specifications of both models in advance, so that $\mc{V}_\mr{S}$, $\mc{V}_\mr{L}$, and the induced common vocabulary $\mc{V}_\mr{C}$ are known prior to generation.
In practice, this can be realized by sharing metadata, such as tokenizer configuration files or model identifiers, in advance.
Note that, since tokenizer specifications are fixed for each model, this does not incur additional communication overhead during generation.

At each verification round $t$, the SLM distribution $p_{t,n}(x)$ is extended to $\mc{V}_\mr{L}$ by suppressing the probability mass outside $\mc{V}_\mr{C}$ and renormalizing:
\begin{align}
	p_{t,n}^\mr{c} (x) =
	\begin{cases}
		\frac{p_{t,n}(x)}{\sum_{x'\in\mc{V}_\mr{C}} p_{t,n}(x')}, & x \in \mc{V}_\mr{C}, \\
		0, & x \in \mc{V}_\mr{O},
	\end{cases}
	\quad \forall x \in \mc{V}_\mr{L}.
\end{align}
Since $p_{t,n}^\mr{c}(x)$ is defined over $\mc{V}_\mr{L}$, it is aligned with $q_{t,n}(x)$ on the same token space.
Therefore, by replacing $p_{t,n}(x)$ with $p_{t,n}^\mr{c}(x)$, the CoSD framework in Section~\ref{subsec:cosd} can be naively extended to heterogeneous vocabulary scenarios, while maintaining the verification and resampling procedures.
Specifically, the acceptance criterion in~\eqref{eq:acceptance} is modified by replacing $p_{t,n}(x)$ with $p_{t,n}^{\mathrm{c}}(x)$, and
the residual distribution in~\eqref{eq:p_res} is redefined over $\mc{V}_\mr{L}$ as
\begin{align}
	p_\mr{res}(x) = \frac{\max(q_{t,n}(x) - p_{t,n}^\mr{c}(x), 0)}{\sum_{x' \in \mc{V}_\mr{L}} \max(q_{t,n}(x') - p_{t,n}^\mr{c}(x'), 0)}, \quad \forall x \in \mc{V}_\mr{L}.
	\label{eq:p_res_het}
\end{align}
Under this extension, a finalized token may lie either in $\mc{V}_\mr{C}$ or in $\mc{V}_\mr{O}$. 
If it lies in $\mc{V}_\mr{C}$, it is directly appended as a single token to the current prefix. If it lies in $\mc{V}_\mr{O}$, the server decodes the token into a string, and the device tokenizes the received string using the on-device SLM tokenizer. The resulting SLM-token sequence is appended to the device prefix for the next-round drafting. 
Note that the decoded string corresponding to an LLM-only token may be tokenized into multiple tokens by the on-device SLM tokenizer. 
The overall process is depicted in Figure~\ref{fig:cosd}.

Nevertheless, this naive extension remains impractical due to prohibitive communication overhead. 
Specifically, whenever a candidate token is rejected, the full LLM distribution over $\mc{V}_\mr{L}$ must be sent to the device for resampling, which incurs substantial downlink latency and degrades token generation speed.
Therefore, the key challenge is to design a communication-efficient CoSD framework for heterogeneous vocabularies that avoids transmitting the full LLM distribution while guaranteeing lossless decoding with respect to the server LLM.

\section{X-CoSD}\label{sec:method}
In this section, we propose X-CoSD, a lossless and communication-efficient CoSD framework for heterogeneous SLM--LLM vocabularies. 
The key design principle of X-CoSD is to redesign the resampling procedure so that exact decoding can be preserved while reducing communication load.
The core of X-CoSD is HR, a device-server hybrid residual resampling scheme that serves as the foundation of our framework.
Building on HR, we propose SR-DV, a variant that further reduces communication load through server-sampled replacement candidates and device-side verification.

\subsection{Hybrid resampling}
To reduce the downlink overhead of transmitting the full LLM distribution in the naive extension while preserving lossless decoding, we leverage a structural property of the residual distribution in~\eqref{eq:p_res_het}. 
For notational simplicity, we focus on a fixed verification round $t$ and a rejected position $n$, and omit the indices $t$ and $n$ in the following unless otherwise stated.

Since $p^\mr{c}(x) =0$ for all $x \in \mc{V}_\mr{O}$, $p_\mr{res}(x)$ can be expressed in a region-wise form as
\begin{align}
	p_\mr{res}(x) = 
	\begin{cases}
		\frac{\max(q(x)-p^\mr{c}(x),0)}{\theta_\mr{c} + \theta_\mr{o}}, & x \in \mc{V}_\mr{C}, \\
		\frac{q(x)}{\theta_\mr{c} + \theta_\mr{o}}, & x \in \mc{V}_\mr{O},
	\end{cases}
	\quad \forall x \in \mc{V}_\mr{L},
	\label{eq:res_decomp}
\end{align}
where $\theta_\mr{c} \triangleq \sum_{x \in \mc{V}_\mr{C}} \max(q(x)-p^\mr{c}(x),0)$ and $\theta_\mr{o} \triangleq \sum_{x \in \mc{V}_\mr{O}} q(x)$ represent the total unnormalized residual mass over $\mc{V}_\mr{C}$ and $\mc{V}_\mr{O}$, respectively.
This decomposition further implies that $p_\mr{res}(x)$ can be expressed as a mixture of two region-wise distributions:
\begin{align}
	p_{\mr{res}}(x)
	= \frac{\theta_\mr{c}}{\theta_\mr{c} + \theta_\mr{o}}  p_{\mr{res}}^\mr{c}(x)
	+ \frac{\theta_\mr{o}}{\theta_\mr{c} + \theta_\mr{o}}  p_{\mr{res}}^\mr{o}(x), \quad \forall x \in \mc{V}_\mr{L},
\end{align}
where $p_{\mr{res}}^\mr{c}(x)$ and $p_{\mr{res}}^\mr{o}(x)$ are the normalized residual distributions over $\mc{V}_\mr{C}$ and $\mc{V}_\mr{O}$, respectively, defined as
\begin{align}
	p_\mr{res}^\mr{c}(x) \triangleq
	\begin{cases} 
		\frac{\max(q(x)-p^\mr{c}(x),0)}{\theta_\mr{c}},& x \in \mc{V}_\mr{C},\\
		0, & x \in \mc{V}_\mr{O},
	\end{cases}
	\quad
	p_\mr{res}^\mr{o}(x) \triangleq
	\begin{cases} 
		0,& x \in \mc{V}_\mr{C},\\
		\frac{q(x)}{\theta_\mr{o}}, & x \in \mc{V}_\mr{O},
	\end{cases}
	\quad \forall x \in \mc{V}_\mr{L}.
\end{align}
This formulation establishes that sampling from $p_{\mr{res}}(x)$ is exactly equivalent to the following two-stage procedure: 
1) selecting a resampling region $G \in \{\mc{V}_\mr{C}, \mc{V}_\mr{O}\}$ with probabilities proportional to $\theta_\mr{c}$ and $\theta_\mr{o}$, and 
2) sampling from the corresponding conditional distribution.

\begin{figure}[t]
	\centering
	\begin{subfigure}[t]{\columnwidth}
		\centering
		\includegraphics[width=0.85\linewidth]{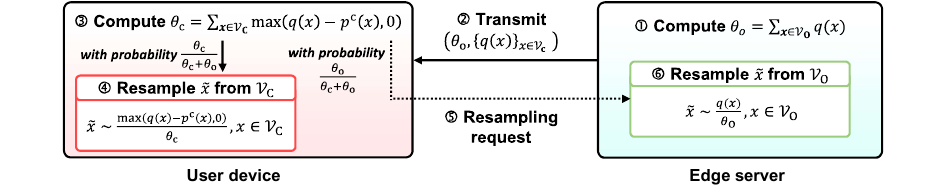}
		\caption{}
		\label{fig:hybrid}
	\end{subfigure}
	\begin{subfigure}[t]{\columnwidth}
		\centering
		\includegraphics[width=0.85\linewidth]{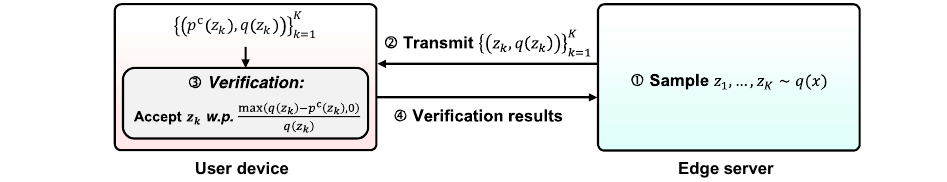}
		\caption{}
		\label{fig:rejection}
	\end{subfigure}
	\caption{Illustrations of (a) HR and (b) SR-DV procedures after rejection occurs.}
\end{figure}

Based on this observation, HR realizes exact residual resampling in a communication-efficient manner, as illustrated in Figure~\ref{fig:hybrid}. 
Instead	 of transmitting the full LLM distribution upon rejection, the server sends only the information required for this two-stage sampling, i.e., the LLM distribution over the common region $\{q(x)\}_{x\in\mc{V}_\mr{C}}$ and the scalar $\theta_\mr{o}$. 
Using this information, the user device computes $\theta_\mr{c}$ and selects the resampling region $G \in \{\mc{V}_\mr{C}, \mc{V}_\mr{O}\}$ according to 
\begin{align}
	G = 
	\begin{cases}
		\mc{V}_\mr{C}, & \textrm{w.p.} \frac{\theta_\mr{c}}{\theta_\mr{c} + \theta_\mr{o}}, \\
		\mc{V}_\mr{O}, & \textrm{w.p.} \frac{\theta_\mr{o}}{\theta_\mr{c} + \theta_\mr{o}}.
	\end{cases}
\end{align}
Given $G$, resampling proceeds as follows.
\begin{itemize}
	\item \textbf{On-device resampling ($G= \mc{V}_\mr{C}$):} The user device locally samples the replacement token $\tilde{x}$ from $p_\mr{res}^\mr{c}(x)$.
	\item \textbf{Server resampling ($G= \mc{V}_\mr{O}$):} The user device requests server-side resampling, and the edge server samples $\tilde{x}$ from $p_\mr{res}^\mr{o}(x)$.
	Since $\tilde{x} \in \mc{V}_\mr{O}$ is an LLM-only token, the server detokenizes $\tilde{x}$ into its corresponding string and sends the string to the device. Then, the device tokenizes the received string with the SLM tokenizer and appends the resulting SLM-token sequence to the current prefix for the next-round drafting. Although the server resampling procedure causes additional communication latency, it is small because the user device sends only a resampling request and the server returns only the detokenized string corresponding to a single LLM-only token, rather than a probability distribution.
\end{itemize}

HR reduces communication overhead by restricting the downlink transmission to the common region $\mc{V}_\mr{C}$, rather than the full LLM distribution.
Moreover, the following theorem establishes that X-CoSD based on HR preserves the server LLM distribution, and thus guarantees lossless decoding.
\begin{theorem}
	The token finalized by \emph{X-CoSD} follows the server LLM distribution $q(x)$.
	\label{thm:x_cosd}
\end{theorem}
The proof is provided in Appendix~\ref{appendix:thm1}.

\subsection{Server resampling with device verification}
In HR, when a candidate token is rejected, the server transmits the LLM distribution over $\mc{V}_{\mr{C}}$ to the device. 
Although this reduces the communication load compared to transmitting the full distribution, it can still incur substantial downlink load. 
To further reduce this overhead, we propose SR-DV, which performs residual resampling using a small number of server-generated replacement candidates and device-side verification, while invoking HR only when all sampled candidates are rejected.

In SR-DV, the server first generates replacement candidates and the device verifies them. 
Specifically, the server samples $K$ replacement candidates $\mb{z}=[z_1,\dots,z_K]$, where each $z_k$ is sampled i.i.d. from $q(x)$, and sends $\mb{z}$ along with their corresponding probabilities $\{q(z_k)\}_{k=1}^K$ to the device. 
The device accepts each candidate $z_k$ with probability $\alpha(z_k)$, where $\alpha(\cdot)$ denotes an acceptance probability function characterized by \textbf{Lemma~\ref{lemma}}.
If any candidate is accepted, it is used as the replacement token. Otherwise, the server generates a new set of $K$ candidates. 
This process is repeated for at most $M$ iterations; if all $KM$ candidates are rejected, the procedure falls back to HR, in which the server transmits the remaining LLM probabilities over $\mc{V}_{\mr{C}}$ that have not already been sent during SR-DV.
The overall process is illustrated in Figure~\ref{fig:rejection}. 
SR-DV is designed to reduce communication latency by probabilistically avoiding the transmission of the common-region distribution required by HR. 
Although SR-DV may introduce additional interactions, each interaction transmits only $K$ replacement candidate tokens and their probabilities; if any candidate is accepted, the costly HR transmission is skipped. 
The maximum number of iterations $M$ bounds the additional interactions and can be chosen according to the target latency budget, while fallback to HR preserves exact residual resampling when all candidates are rejected. 
The following proposition provides the expected number of SR-DV iterations and the expected latency per verification round, which includes on-device drafting, server verification, and resampling upon rejection.
\begin{proposition}
	Let $I$ denote the number of iterations performed in \emph{SR-DV}. Then, 
	\begin{align}
		\mathbb{E}[I] = \frac{1 - (1-\eta)^M}{\eta},
	\end{align}
	where $\eta \triangleq 1 - \left(1-\sum_{x' \in \mc{V}_\mr{L}}\alpha(x')q(x')\right)^K$ is the probability that at least one replacement candidate is accepted in one \emph{SR-DV} iteration.
	
	 Given a length-$N$ drafted token sequence, we denote $T$ as the latency per verification round. Let $T_\mr{S}$,  $T_\mr{L}$, $T_\mr{up}$, $T_\mr{down}^\mr{acc}$, $T_\mr{down}^\mr{SRDV}$, and $T_\mr{down}^\mr{HR}$ denote the single forward pass latency of the on-device SLM, the single forward pass latency of the server LLM, uplink latency, downlink latency for the full acceptance case, downlink latency for one \emph{SR-DV} iteration, and downlink latency for \emph{HR} fallback, respectively. Assuming that the token-wise acceptance events are i.i.d., the full acceptance probability of the length-$N$ candidate sequence is given by $\Lambda \triangleq \left(\sum_{x'\in\mc{V}_\mr{C}} \min\left(p^\mr{c}(x'), q(x')\right)\right)^N$. Then, 
	\begin{align}
		\mathbb{E}[T]
		&= NT_\mr{S}+T_\mr{L}+T_\mr{up} +\Lambda T_\mr{down}^\mr{acc} +(1-\Lambda)\left( \mathbb{E}[I]T_\mr{down}^\mr{SRDV} +(1-\eta)^M T_\mr{down}^\mr{HR}\right).
	\end{align}
\end{proposition}
The proof is provided in Appendix~\ref{appendix:prop}.

To preserve lossless decoding, SR-DV must ensure that the generated token follows $p_{\mr{res}}(x)$ in~\eqref{eq:p_res_het}. 
The following lemma characterizes the acceptance probability function $\alpha(x)$ that ensures this property.
\begin{lemma}\label{lemma}
	The acceptance probability function $\alpha(x)$ that ensures the token produced by \emph{SR-DV} follows the residual distribution $p_\mr{res}(x)$ is given by
	\begin{align}
		\alpha(x)=\frac{\max(q(x)-p^{\mr{c}}(x),0)}{q(x)}.
		\label{eq:acc_SR-DV}
	\end{align}
	Among feasible acceptance functions satisfying $0 \leq \alpha(x) \leq 1$, this choice maximizes the probability that a replacement candidate is accepted and thus minimizes the fallback probability to HR.
\end{lemma}
The proof is provided in Appendix~\ref{appendix:lemma}. 
Note that replacement candidates in $\mc{V}_\mr{O}$ are accepted with probability one, since $\alpha(x)=1$ for $x \in \mc{V}_\mr{O}$ in~\eqref{eq:acc_SR-DV}.

SR-DV reduces communication overhead since the common-region distribution is transmitted only in the worst case where all $KM$ candidates are rejected and the procedure falls back to HR. 
We refer to the X-CoSD variant enabled by SR-DV as X-CoSD-E.
Moreover, the following theorem establishes that X-CoSD-E preserves the server LLM distribution and thus guarantees lossless decoding.
\begin{theorem}
	The token finalized by \emph{X-CoSD-E} follows the server LLM distribution $q(x)$.
\end{theorem}
The proof is provided in Appendix~\ref{appendix:thm2}.

\section{Experiments}\label{sec:experiments}
In this section, we evaluate the proposed X-CoSD and X-CoSD-E frameworks.
We consider a system where the user device and the server are equipped with an NVIDIA TITAN RTX and an NVIDIA A100 (40GB), respectively.
The experimental setup is described as follows.

\paragraph{Models and Datasets}
To establish a heterogeneous-vocabulary setting, we use Vicuna-68M~\cite{yang24} as the on-device SLM and consider  Llama-3.1-8B~\cite{grattafiori24} and Qwen2-7B~\cite{yang24_qwen} as the server LLMs.
We use the pretrained Hugging Face models in FP16 precision without modification. Detailed model identifiers are provided in Appendix~\ref{appendix:implementation}.
During the experiments, the softmax temperature of all models is set to $0.3$, and the maximum generation length is set to $128$. 
We evaluate the proposed and baseline methods in terms of generation speed and response quality across diverse tasks.
Specifically, we use WMT-DeEn~\cite{bojar14} for machine translation, XSum~\cite{narayan18} and CNN/DailyMail~\cite{hermann15} for summarization, GSM8K~\cite{cobbe21} for mathematical reasoning, and MMLU~\cite{hendrycks20} for question answering.
\paragraph{Baselines} We consider the following baselines for comparison.
\begin{itemize}
	\item \textbf{Naive:} A TLI \cite{timor25}-based naive extension of CoSD to the heterogeneous vocabulary setting, described in Section~\ref{sec:naive}.
	\item \textbf{Uplink (UL):} The user device always transmits the candidate tokens and corresponding distributions over $\mc{V}_\mr{C}$ to the server. Upon rejection, the server performs resampling using $p_\mr{res}(x)$ in~\eqref{eq:p_res_het}.
	\item \textbf{Greedy resampling (GR):} A communication-efficient but lossy variant of X-CoSD. When rejection occurs, the server replaces the rejected token with the top-1 token from the server LLM distribution. GR is used for generation quality comparison.
	\item \textbf{Target resampling (TR):} Similar to GR, upon rejection, the server directly resamples a new token from the server LLM distribution. TR is also used for generation quality comparison.
	\item \textbf{U-HLM \cite{oh24}:} We adapt U-HLM~\cite{oh24} to the heterogeneous vocabulary setting. 
	The user device requests verification only for candidate tokens whose uncertainty exceeds a predefined threshold $u_\mr{th}$. 
	When verification is requested, the device transmits the SLM distribution over $\mc{V}_\mr{C}$. Upon rejection, the server resamples a new token from $p_\mr{res}(x)$ in~\eqref{eq:p_res_het}.
	\item \textbf{Server LLM:} The user device uploads the prompt to the server, and the server LLM autoregressively generates the output.
\end{itemize}
Throughout the experiments, we report the average performance over three runs, where each run is conducted over 1,000 samples per dataset.
For X-CoSD-E, we set $K=20$ and $M=10$.
More details on experimental setup are in~Appendix~\ref{appendix:implementation}.

In Table~\ref{tab:score}, we report the relative generation quality of the proposed and baseline methods with respect to the server LLM.
Following~\cite{jeon24}, generation quality scores are evaluated using BLEU for WMT-DeEn, ROUGE-2 for XSum and CNN/DailyMail. 
The scores for GSM8K and MMLU are measured by accuracy.
Since Naive and UL differ only in the resampling location, their final output quality is identical; thus, we report only the performance of the Naive method.
The on-device SLM yields substantially lower generation quality than the server LLM due to its limited model capacity, which underscores the need for collaborative LLM inference.
X-CoSD and X-CoSD-E achieve generation quality comparable to the server LLM, consistent with their lossless guarantees. 
For the same reason, Naive and UL also attain similar quality.
Small deviations from 1.0 are expected due to sampling variability and finite-sample evaluation.
In contrast, the heuristic resampling methods GR and TR show unstable generation quality since they do not preserve the server LLM distribution.
U-HLM also suffers from severe quality degradation due to skipped verification.
These results highlight the importance of lossless CoSD methods for heterogeneous vocabularies.

\begin{table}[!t]
	\centering
	\caption{Relative scores compared to Server LLM}
	{\footnotesize
		\begin{tabular}{llccccc}
			\toprule
			LLM type & Method & WMT-DeEn & XSum & CNN/DailyMail & GSM8K & MMLU \\
			\midrule
			\multirow{7}{*}{Qwen2-7B} & Server LLM & 1.0000 & 1.0000 & 1.0000 & 1.0000 & 1.0000\\
			\cmidrule(lr){2-7}
			& \textbf{X-CoSD (prop.)} & \textbf{1.0471} & \textbf{1.0248} & \textbf{0.9964} & \textbf{1.0048} & \textbf{0.9925} \\
			& \textbf{X-CoSD-E (prop.)} & \textbf{1.0239} & \textbf{0.9832} & \textbf{1.0031} & \textbf{0.9979} & \textbf{0.9925} \\
			& Naive & 1.0361 & 0.9960 &0.9910 & 1.0158 & 0.9785 \\
			& GR & 1.0938 & 1.0197 & 0.7836 & 1.0042 & 0.9828 \\
			& TR & 1.0239 & 1.0086 & 0.7850 & 0.9895 & 0.9462 \\
			& U-HLM ($u_\mr{th}=0.5$)& 0.3289 & 0.6709 & 0.8697 & 0.0622 & 0.6437 \\
			\midrule
			\multirow{7}{*}{Llama-3.1-8B} & Server LLM & 1.0000 & 1.0000 & 1.0000 & 1.0000 & 1.0000\\
			\cmidrule(lr){2-7}
			& \textbf{X-CoSD (prop.)} & \textbf{1.0217} & \textbf{0.9676} & \textbf{0.9906} & \textbf{1.0021} & \textbf{1.0326} \\
			& \textbf{X-CoSD-E (prop.)} & \textbf{1.0169} & \textbf{0.9794} & \textbf{0.9985} & \textbf{0.9986} & \textbf{1.0240} \\
			& Naive & 1.0035 & 0.9842 & 0.9987 & 0.9944 & 1.0137 \\
			& GR & 1.1057 & 0.9498 & 0.9719 & 0.9916& 0.9631\\
			& TR & 0.9372 & 0.8471& 0.9890 & 0.9783& 0.9528\\
			& U-HLM ($u_\mr{th}=0.5$)& 0.3570 & 0.6006 & 0.7955 & 0.0883 & 0.4481\\
			\midrule
			& On-device SLM & 0.0489 & 0.2854 & 0.2305 & 0.0327 & 0.4682\\
			\bottomrule
		\end{tabular}
	}
	\label{tab:score}
\end{table}

\begin{table}[!t]
	\centering
	\caption{Communication load per generated token for $N=2$ (bits)}
	{\footnotesize
		\begin{tabular}{llcccc}
			\toprule
			\multirow{2}{*}{LLM type} & \multirow{2}{*}{Method} & \multicolumn{2}{c}{WMT-DeEn} & \multicolumn{2}{c}{XSum}\\
			\cmidrule(lr){3-4}
			\cmidrule(lr){5-6}
			& & Uplink & Downlink & Uplink & Downlink \\
			\midrule
			\multirow{5}{*}{Qwen2-7B} & \textbf{X-CoSD (prop.)} & \textbf{40.4} & \textbf{0.2 M} & \textbf{112.5} & \textbf{0.2 M}\\
			& \textbf{X-CoSD-E (prop.)} & \textbf{47.3} & \textbf{384.2} & \textbf{107.0} & \textbf{369.4}\\
			& Naive & 32.6 & 1.2 M & 106.8 & 1.2 M \\
			& UL & 0.2 M & 11.3 & 0.2 M & 11.2 \\
			& U-HLM ($u_\mr{th}=0.5$)& 0.1 M & 6.7 & 0.1 M & 5.6 \\
			\midrule
			\multirow{5}{*}{Llama-3.1-8B} & \textbf{X-CoSD (prop.)} & \textbf{52.7} & \textbf{0.2 M} & \textbf{60.1} & \textbf{0.1 M} \\
			& \textbf{X-CoSD-E (prop.)} & \textbf{44.1} & \textbf{378.7} & \textbf{56.8} & \textbf{265.8} \\
			& Naive & 44.4 & 1.1 M  & 57.6 & 0.7 M \\
			& UL & 0.2 M & 11.1 & 0.2 M & 9.0\\
			& U-HLM ($u_\mr{th}=0.5$)& 0.1 M & 6.5 & 0.1 M &  4.5\\
			\bottomrule
		\end{tabular}
	}
	\label{tab:load_per_token}
\end{table}
In Table~\ref{tab:load_per_token}, we report the uplink and downlink communication load per generated token, measured in bits.
We exclude the server LLM, GR, and TR from this comparison, since they do not require distribution transmission and therefore incur negligible communication overhead.
This quantity is computed as the total communication load divided by the total number of generated tokens.
Due to space limitations, we present the results on WMT-DeEn and XSum here, while the results on the other datasets are provided in Appendix~\ref{appendix:load}.
As expected, X-CoSD substantially reduces the downlink communication load compared to Naive by restricting downlink transmission, through HR, to the common-vocabulary region.
Building on this, X-CoSD-E further reduces the downlink communication load through SR-DV, which falls back to HR only when all replacement candidates are rejected, thereby significantly improving communication efficiency for practical deployment.
In contrast, UL incurs substantial uplink communication load, making it less suitable for practical collaborative scenarios where the uplink is typically the bottleneck.
Although U-HLM reduces the uplink communication load relative to UL by verifying only uncertain candidate tokens, it still requires uplink transmission of the SLM distribution for those tokens and does not guarantee lossless decoding.
Overall, these results underscore the practical advantage of X-CoSD and X-CoSD-E in achieving lossless decoding while substantially reducing communication overhead.

\begin{figure}[!t]
	\centering
	\begin{subfigure}[t]{0.497\columnwidth}
		\centering
		\includegraphics[width=0.494\linewidth]{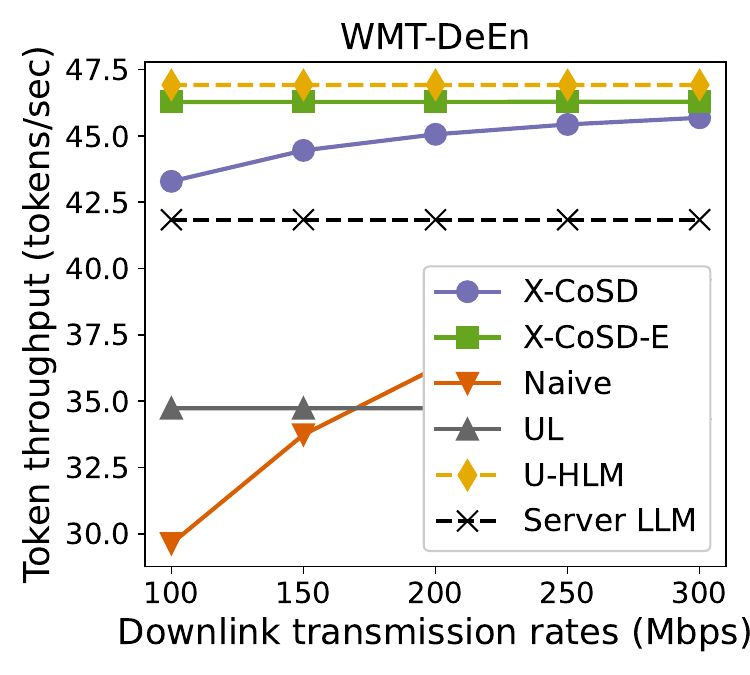}
		\includegraphics[width=0.494\linewidth]{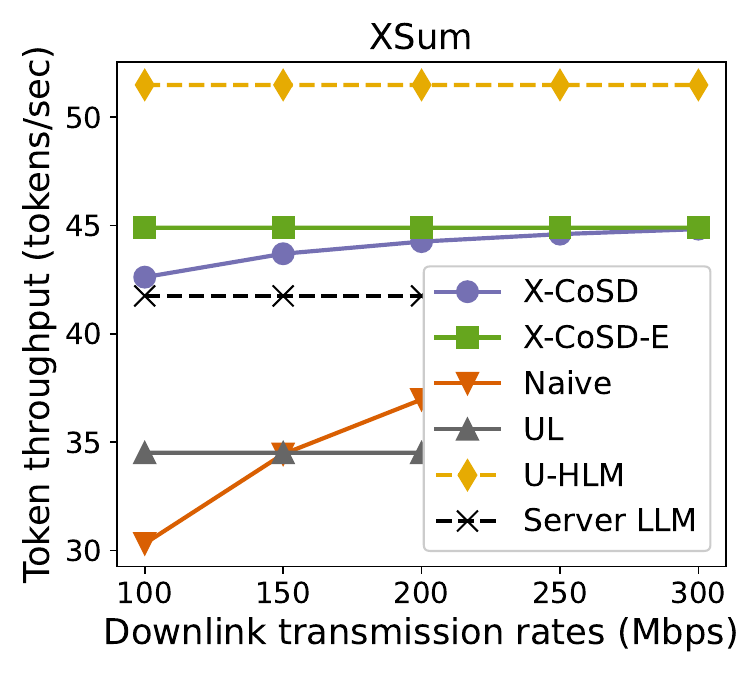}
		\caption{Qwen2-7B}
		\label{fig:qwen_pair}
	\end{subfigure}
	\centering
	\begin{subfigure}[t]{0.497\columnwidth}
		\centering
		\includegraphics[width=0.494\linewidth]{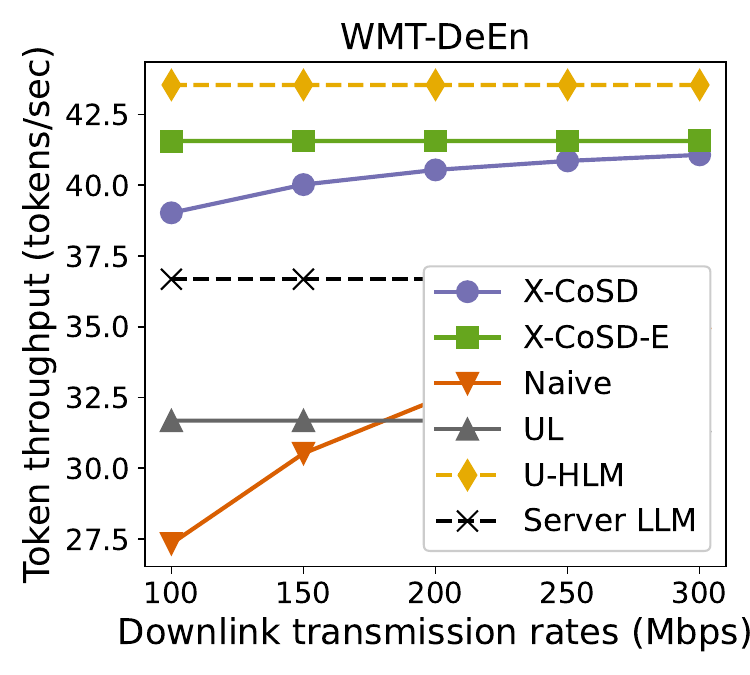}
		\includegraphics[width=0.494\linewidth]{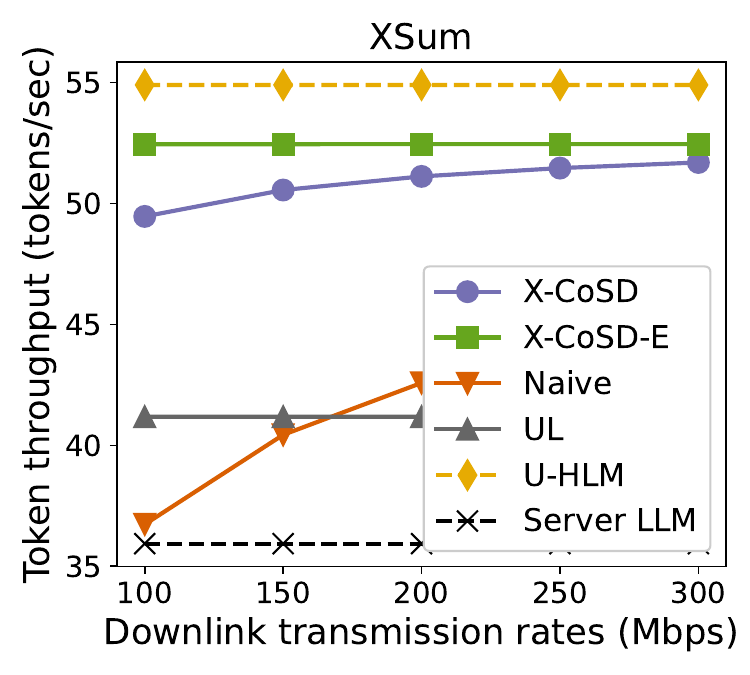}
		\caption{Llama-3.1-8B}
		\label{fig:llama_pair}
	\end{subfigure}
	\caption{Token throughput versus the downlink transmission rates.}
	\label{fig:sim_tput_wmt_xsum}
\end{figure}

Figure~\ref{fig:sim_tput_wmt_xsum} shows token throughput, measured in generated tokens per second, against the downlink transmission rate for WMT-DeEn and XSum.
For each method, we report the best token throughput over candidate sequence lengths $N\in \{1,\dots,8\}$, a practical range commonly used in conventional SD~\cite{leviathan23, chen23, miao24, ramakrishnan25}. 
Based on real-world mid-band 5G measurements~\cite{rochman25}, we set the uplink transmission rate to 30 Mbps and vary the downlink transmission rate from 100 to 300 Mbps.
As shown in the figure, X-CoSD and X-CoSD-E consistently outperform Naive, owing to their reduced downlink communication load, and both methods achieve higher token throughput than server LLM inference.
However, similar to Naive, the token throughput of X-CoSD remains heavily dependent on the downlink transmission rate, since it still requires transmission of the LLM distribution over $\mc{V}_\mr{C}$ whenever rejection occurs.
In contrast, by reducing communication load through SR-DV, X-CoSD-E improves upon X-CoSD and maintains high token throughput across the entire range of downlink transmission rates, highlighting its practical advantage in communication-limited regimes.
UL shows poor token throughput, since its substantial uplink communication load is unfavorable in practical wireless networks, where the uplink transmission rate is typically limited.
Although U-HLM can achieve the highest token throughput, it does not guarantee lossless decoding.
Results on CNN/DailyMail, GSM8K, and MMLU are provided in Appendix~\ref{appendix:tput} and show consistent trends.

\begin{figure}[!t]
	\centering
	\begin{subfigure}[t]{0.497\columnwidth}
		\centering
		\includegraphics[width=1.0\linewidth]{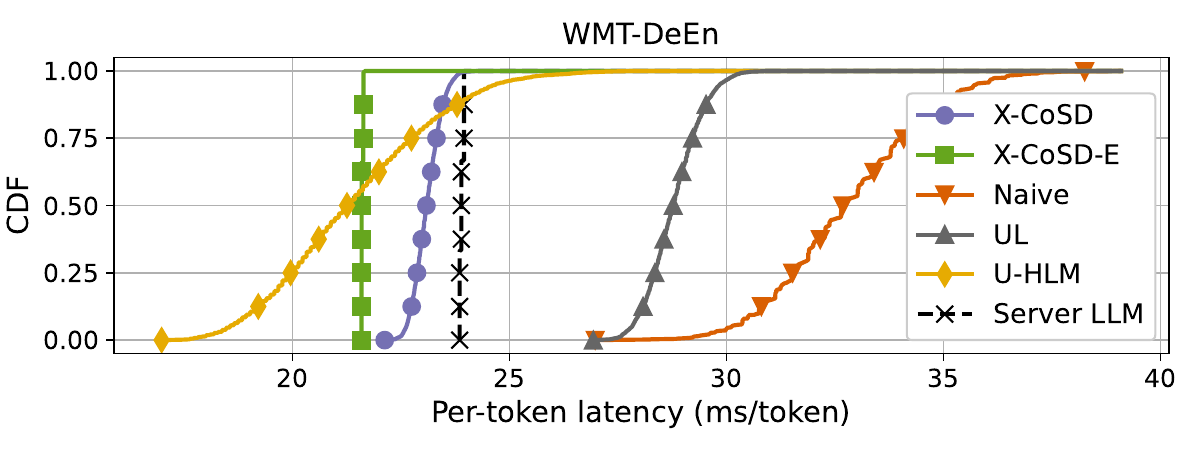}
		\caption{Qwen2-7B}
		\label{fig:cdf_qwen}
	\end{subfigure}
	\centering
	\begin{subfigure}[t]{0.497\columnwidth}
		\centering
		\includegraphics[width=1.0\linewidth]{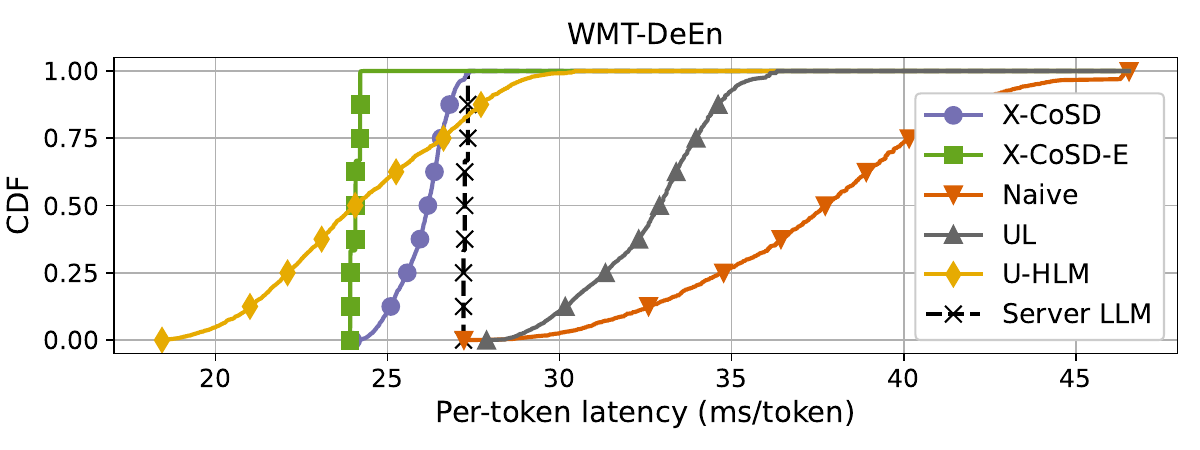}
		\caption{Llama-3.1-8B}
		\label{fig:cdf_llama}
	\end{subfigure}
	\caption{CDF of per-token latency.}
	\label{fig:sim_cdf_wmt}
\end{figure}

Figure~\ref{fig:sim_cdf_wmt} shows the CDF of per-token latency on WMT-DeEn, where per-token latency is computed as the latency divided by the number of generated tokens for each sample. We set the uplink and downlink transmission rates to 30 Mbps and 100 Mbps, respectively. X-CoSD and X-CoSD-E achieve lower per-token latency than the lossless baselines, demonstrating their effectiveness. In particular, X-CoSD-E consistently outperforms X-CoSD in terms of per-token latency owing to the communication load reduction achieved by the SR-DV mechanism. Although U-HLM shows slightly lower per-token latency, it is a lossy baseline and does not preserve the server LLM distribution. Results on XSum, CNN/DailyMail, GSM8K, and MMLU also show consistent trends and are provided in Appendix~\ref{appendix:cdf}.

\section{Conclusions}
In this work, we presented X-CoSD and X-CoSD-E, lossless and communication-efficient CoSD frameworks for heterogeneous vocabularies between the on-device SLM and the server LLM.
We showed that they preserve the server LLM distribution through HR and SR-DV, respectively, while X-CoSD-E further improves communication efficiency.
Experiments demonstrated that the proposed methods improve token throughput without sacrificing the generation quality of the server LLM.
Together, these frameworks represent a practical step toward real-world CoSD deployment under heterogeneous SLM--LLM vocabularies and limited communication resources.

\bibliographystyle{unsrtnat}
\bibliography{ref}

\newpage
\appendix

\section{Proofs}\label{appendix:proof}
\paragraph{Basic notations}
\begin{itemize}
	\item $X$: Candidate token generated by the on-device SLM.
	\item $Y$: Final output token of X-CoSD(-E).
	\item $\ms{A}$: Event that $X$ is accepted.
	\item $\ms{R}(=\ms{A}^c)$: Event that $X$ is rejected.
	\item $p^\mr{c}(x)$: On-device SLM distribution truncated to $\mc{V}_\mr{C}$.
	\item $q(x)$: Server LLM distribution.	
	\item $p_\mr{res}(x) = \frac{\max(q(x)-p^\mr{c}(x),0)}{\sum_{x'\in\mc{V}_\mr{L}} \max(q(x')-p^\mr{c}(x'),0)}$: Residual distribution.
\end{itemize}
\subsection{Proof of Theorem~\ref{thm:x_cosd}} \label{appendix:thm1}
By the law of total probability, $\forall x \in \mc{V}_\mr{L}$, we have
\begin{align}
	\prx{Y=x} = \prx{Y=x,\ms{A}} + \prx{Y=x,\ms{R}}.
	\label{eq:hr_total}
\end{align}
\paragraph{1) Computing $\prx{Y=x,\ms{A}}$ in~\eqref{eq:hr_total}} Under acceptance, the final output token equals the candidate token. Thus, we can rewrite $\prx{Y=x, \ms{A}}$ as
\begin{align}
	\prx{Y=x, \ms{A}} &= \prx{X=x, \ms{A}}\\
	&= \prx{\ms{A} \mid X=x} \prx{X=x} \\
	&= \min\left(1,\frac{q(x)}{p^\mr{c}(x)}\right) p^\mr{c}(x)\\
	&= \min(p^\mr{c}(x), q(x)).
	\label{eq:hr_term1}
\end{align}
\paragraph{2) Computing $\prx{Y=x,\ms{R}}$ in~\eqref{eq:hr_total}} We can rewrite $\prx{Y=x,\ms{R}}$ as 
\begin{align}
	\prx{Y=x, \ms{R}} = \prx{Y=x \mid \ms{R}}\prx{\ms{R}}\label{eq:hr_rej}
\end{align}
First, using the law of total probability, $\prx{Y=x \mid \ms{R}}$ in~\eqref{eq:hr_rej} can be represented as 
\begin{align}
	\prx{Y=x \mid \mathsf{R}} &= \underbrace{\prx{Y=x, G = \mc{V}_{\mr{C}} \mid \mathsf{R}}}_{\triangleq a_1} + \underbrace{\prx{Y=x, G =  \mc{V}_{\mr{O}} \mid \mathsf{R}}}_{\triangleq a_2},
	\label{eq:hr_region}
\end{align}
Recall that, under HR, the resampling region $G\in\{\mc{V}_\mr{C}, \mc{V}_\mr{O}\}$ upon rejection is selected as
\begin{align}
	G = 
	\begin{cases}
		\mc{V}_\mr{C}, & \textrm{w.p.} \frac{\theta_\mr{c}}{\theta_\mr{c} + \theta_\mr{o}},\\
		\mc{V}_\mr{O}, & \textrm{w.p.} \frac{\theta_\mr{o}}{\theta_\mr{c} + \theta_\mr{o}},
	\end{cases}
\end{align}
where $\theta_\mr{c} = \sum_{x \in \mc{V}_\mr{C}} \max(q(x)-p^\mr{c}(x),0)$ and $\theta_\mr{o} = \sum_{x \in \mc{V}_\mr{O}} q(x)$. 
Given the selected region $G$, the replacement token is sampled from the corresponding conditional residual distribution.
Specifically, if $G=\mc{V}_\mr{C}$, the token is drawn from $p_\mr{res}^\mr{c}(x)$, whereas if $G=\mc{V}_\mr{O}$, it is drawn from $p_\mr{res}^\mr{o}(x)$, defined as
\begin{align}
	p_\mr{res}^\mr{c}(x) =
	\begin{cases} \frac{\max(q(x)-p^\mr{c}(x),0)}{\theta_\mr{c}},& x \in \mc{V}_\mr{C},\\
		0, & x \in \mc{V}_\mr{O},
	\end{cases}
\end{align}
\begin{align}
	p_\mr{res}^\mr{o}(x) =
	\begin{cases} 
		0,& x \in \mc{V}_\mr{C},\\
		\frac{q(x)}{\theta_\mr{o}}, & x \in \mc{V}_\mr{O}.
	\end{cases}
\end{align}
Therefore, $a_1$ and $a_2$ in~\eqref{eq:hr_region} are obtained as follows.
\begin{align}
	a_1 &= \prx{Y=x \mid G = \mc{V}_{\mr{C}}, \mathsf{R}} \prx{G = \mc{V}_{\mr{C}} \mid \mathsf{R}} \\
	&= p_\mr{res}^\mr{c}(x) \frac{\theta_\mr{c}}{\theta_\mr{c} + \theta_\mr{o}}\\
	&=
	\begin{cases}
		\frac{\max(q(x)-p^\mr{c}(x),0)}{\theta_\mr{c} + \theta_\mr{o}}, & x \in \mc{V}_\mr{C}, \\
		0, & x \in \mc{V}_\mr{O},
	\end{cases}
	\label{eq:hr_a1}
\end{align}
\begin{align}
	a_2 &= \prx{Y=x \mid G = \mc{V}_{\mr{O}}, \mathsf{R}} \prx{G = \mc{V}_{\mr{O}} \mid \mathsf{R}} \\
	&= p_\mr{res}^\mr{o}(x) \frac{\theta_\mr{o}}{\theta_\mr{c} + \theta_\mr{o}}\\
	&=
	\begin{cases}
		0, & x \in \mc{V}_\mr{C}, \\
		\frac{q(x)}{\theta_\mr{c} + \theta_\mr{o}}, & x \in \mc{V}_\mr{O}.
	\end{cases}
	\label{eq:hr_a2}
\end{align}
By substituting~\eqref{eq:hr_a1} and~\eqref{eq:hr_a2} into~\eqref{eq:hr_region}, we obtain
\begin{align}
	\prx{Y=x \mid \ms{R}} 
	&= \begin{cases}
		\frac{\max(q(x)-p^\mr{c}(x),0)}{\theta_\mr{c} + \theta_\mr{o}}, & x \in \mc{V}_\mr{C}, \\
		\frac{q(x)}{\theta_\mr{c} + \theta_\mr{o}}, & x \in \mc{V}_\mr{O},
		\end{cases}\\
	&= \frac{\max(q(x)-p^\mr{c}(x),0)}{\theta_\mr{c} + \theta_\mr{o}} \label{eq:hr_condition_r_1}\\
	&= p_\mr{res}(x),
	\label{eq:hr_condition_r_2}
\end{align}
where~\eqref{eq:hr_condition_r_1} follows from the fact that $p^\mr{c}(x)=0$ for $x\in\mc{V}_\mr{O}$, which implies $q(x)=\max(q(x)-p^\mr{c}(x),0)$ for $x \in \mc{V}_\mr{O}$.

Next, $\prx{\ms{R}}$ in~\eqref{eq:hr_rej} is given by
\begin{align}
	\prx{\ms{R}} &= \sum_{x \in \mc{V}_\mr{L}} \prx{\ms{R}, X=x} \\
	&= \sum_{x \in \mc{V}_\mr{L}} \prx{\ms{R}\mid X=x}\prx{X=x}\\
	&= \sum_{x \in \mc{V}_\mr{L}} \left(1 - \min\left(1,\frac{q(x)}{p^\mr{c}(x)}\right) \right) p^\mr{c}(x)\\
	&= \sum_{x \in \mc{V}_\mr{L}} \left(p^\mr{c}(x) - \min(p^\mr{c}(x),q(x))\right) \\
	&= \sum_{x \in \mc{V}_\mr{L}} \left(q(x) - \min(p^\mr{c}(x),q(x))\right)  \label{eq:hr_sum1}\\
	&= \sum_{x \in \mc{V}_\mr{L}} \max(q(x)-p^\mr{c}(x),0),
	\label{eq:hr_r}
\end{align}
where~\eqref{eq:hr_sum1} holds due to $\sum_{x\in\mc{V}_\mr{L}} p^\mr{c}(x) = \sum_{x\in\mc{V}_\mr{L}} q(x) = 1$.

Using~\eqref{eq:hr_condition_r_2},~\eqref{eq:hr_r}, and the definition of $p_\mr{res}(x)$, we can rewrite~\eqref{eq:hr_rej} as
\begin{align}
	\prx{Y=x, \ms{R}} &= p_\mr{res}(x) \sum_{x' \in \mc{V}_\mr{L}} \max(q(x')-p^\mr{c}(x'),0).\\
	&= \max(q(x)-p^\mr{c}(x),0).
	\label{eq:hr_term2}
\end{align}

\paragraph{3) Combining $\prx{Y=x,\ms{A}}$ and $\prx{Y=x,\ms{R}}$}
Substituting~\eqref{eq:hr_term1} and~\eqref{eq:hr_term2} into~\eqref{eq:hr_total}, we obtain
\begin{align}
	\prx{Y=x} &= \min(p^\mr{c}(x), q(x)) + \max(q(x)-p^\mr{c}(x),0)\\
	&= q(x),
\end{align}
where the last equality comes from $\min(a,b) + \max(b-a,0)=b$.
This completes the proof.

\subsection{Proof of Proposition~1}\label{appendix:prop}
Conditioned on the rejection event $\ms{R}$, the server generates $K$ replacement candidates $Z_1,\dots,Z_K$ independently from the server LLM distribution $q(x)$.

\paragraph{1) Computing $\mathbb{E}[I]$}
Since each replacement candidate is accepted with probability $\Gamma \triangleq \sum_{x' \in \mc{V}_\mr{L}} \alpha(x')q(x')$, the probability that at least one replacement candidate is accepted in each SR-DV iteration is  
\begin{align}
	\eta \triangleq 1-(1-\Gamma)^K.
\end{align}
Therefore, $I$ is a geometric random variable with success probability $\eta$, capped by the maximum number of iterations $M$.
Accordingly, we can obtain $\mathbb{E}[I]$ as follows.
\begin{align}
	\mathbb{E}[I] 
	&= \sum_{i=1}^{M-1} i (1-\eta)^{i-1} \eta + M (1-\eta)^{M-1} \\
	&= \eta \sum_{i=1}^{M-1} i (1-\eta)^{i-1} + M (1-\eta)^{M-1} \\
	&= \eta \frac{1-M(1-\eta)^{M-1} + (M-1)(1-\eta)^M}{\eta^2} + M (1-\eta)^{M-1} \label{eq:prop_sum}\\
	&= \frac{1 - (1-\eta)^M}{\eta},
\end{align}
where~\eqref{eq:prop_sum} comes from $\sum_{i=1}^{M-1} i \xi^{i-1} = \frac{1-M\xi^{M-1} + (M-1)\xi^M}{(1-\xi)^2}$.

\paragraph{2) Computing $\mathbb{E}[T]$}
 Given the candidate sequence length $N$, let $T_\mr{S}$ and $T_\mr{L}$ denote the latencies of a single forward pass of the on-device SLM and the server-side LLM, respectively.
To derive the full acceptance probability of a length-$N$ candidate sequence, we assume that token-wise acceptance events are i.i.d.
Then, the full acceptance probability $\Lambda$ can be computed as
\begin{align}
	\Lambda \triangleq \left(\sum_{x' \in \mc{V}_\mr{C}} \min(p^\mr{c}(x'), q(x'))\right)^N.
\end{align}
 The uplink communication latency incurred by sending the $N$ candidate tokens and their corresponding probabilities is  
\begin{align}
     T_\mr{up} \triangleq \frac{N(\lceil\log_2|\mc{V}_\mr{C}|\rceil + o_\mr{prob})}{C_\mr{up}},
\end{align}
 where $o_\mr{prob}$ is the number of bits used to represent one probability value, i.e., 16 for FP16 and 32 for FP32, and $C_\mr{up}$ is the uplink transmission rate.
 The downlink latency for the full acceptance case is
\begin{align}
	T_\mr{down}^\mr{acc} \triangleq \frac{\lceil\log_2|\mc{V}_\mr{L}|\rceil}{C_\mr{down}},
\end{align}
where $C_\mr{down}$ denotes the downlink transmission rate. This latency corresponds to the transmission latency of the bonus token ID.
 Upon a rejection event $\ms{R}$, each SR-DV iteration requires the server to send $K$ replacement candidate token IDs and their corresponding probabilities.
The resulting downlink latency per SR-DV iteration is
\begin{align}
	T_\mr{down}^\mr{SRDV} \triangleq \frac{K(\lceil\log_2|\mc{V}_\mr{L}|\rceil + o_\mr{prob})}{C_\mr{down}}.
\end{align}
If all $M$ SR-DV iterations fail, which occurs with probability $(1-\eta)^M$, the procedure falls back to HR. 
 For tractable analysis, we assume $KM$ replacement candidates are distinct. Under this assumption, the downlink latency for transmitting the remaining probability values over the common-vocabulary region required by HR is
\begin{align}
	T_\mr{down}^\mr{HR}
	\triangleq
	\frac{(|\mc{V}_\mr{C}|-KM)o_\mr{prob}}{C_\mr{down}}.
\end{align}
Then, $\mathbb{E}[T]$ is obtained as 
\begin{align}
	\mathbb{E}[T] 
	&= NT_\mr{S} + T_\mr{L} + T_\mr{up} + \Lambda T_\mr{down}^\mr{acc}  + (1-\Lambda) \left(\mathbb{E}[I] T_\mr{down}^\mr{SRDV} + (1-\eta)^M T_\mr{down}^\mr{HR}\right), 
\end{align}
which completes the proof.  

\subsection{Proof of Lemma~1}\label{appendix:lemma}
Let $\ms{S}_k$ denote the event that the $k$-th replacement candidate $Z_k$ is accepted, and let  $\ms{F}_k$  denote the event that $Z_k$ is rejected.
We also define $\Omega$ as the event that at least one of the $K$ candidates is accepted, i.e., $\Omega \triangleq \bigcup_{k=1}^K \ms{S}_k$.
For simplicity, it suffices to consider the single-iteration case $M=1$. 
The general case follows identically by viewing the $KM$ server-sampled candidates across $M$ iterations as a single ordered list of i.i.d.\ samples from $q(x)$ before the fallback to HR.

Our goal is to find an acceptance probability function $\alpha(x)$ that ensures
\begin{align}
	\prx{Y=x\mid\ms{R}} = p_\mr{res}(x). 
\end{align}
By the law of total probability, we have
\begin{align}
	\prx{Y=x\mid\ms{R}} = \prx{Y=x, \Omega\mid\ms{R}} + \prx{Y=x, \Omega^c\mid\ms{R}}.
	\label{eq:lemma_total}
\end{align}

\paragraph{1) Computing $\prx{Y=x, \Omega \mid \ms{R}}$ in~\eqref{eq:lemma_total}}
Let $\ms{F}_{<k} \triangleq \bigcap_{i=1}^{k-1} \ms{F}_i$ denote the event that all candidates before the $k$-th one are rejected.
Conditioning on the event that the $k$-th candidate is the first accepted one, we obtain 
\begin{align}
	\prx{Y=x, \Omega \mid \ms{R}} 
	&= \sum_{k=1}^K \prx{\ms{S}_k, Z_k=x, \ms{F}_{<k} \mid \ms{R}},\\
	&= \sum_{k=1}^K \underbrace{\prx{\ms{S}_k \mid Z_k=x, \ms{F}_{<k}, \ms{R}}}_{\triangleq b_1} \underbrace{\prx{Z_k=x \mid \ms{F}_{<k}, \ms{R}}}_{\triangleq b_2} \underbrace{\prx{\ms{F}_{<k} \mid \ms{R}}}_{\triangleq b_3}.
	\label{eq:lemma_accept_first}
\end{align}

Now, we derive $b_1$, $b_2$, and $b_3$ as follows.
\begin{itemize}
	\item  $b_1$: Conditioned on $\ms{R}$, the candidates $Z_1, \dots, Z_K$ are generated independently from $q(x)$, and the acceptance decision for each candidate is independent of all other candidates and their verification outcomes.
	Hence, we have
	\begin{align}
		b_1 &= \prx{\ms{S}_k \mid Z_k=x, \ms{R}}\\
		&= \alpha(x).\label{eq:lemma_a1}
	\end{align}
	\item $b_2$: Similarly, due to independence,
	\begin{align}
		b_2 &= \prx{Z_k=x \mid \ms{R}}\\
		&= q(x).\label{eq:lemma_a2}
	\end{align}
	\item $b_3$: Let $Z_{<k} \triangleq [Z_1, \dots, Z_{k-1}]$ and let $z_{<k} \triangleq [z_1,\dots,z_{k-1}]$ denote its realization. 
	Then, we can expand $b_3$ as follows.
	\begin{align}
		b_3 &= \sum_{z_1 \in \mc{V}_\mr{L}} \sum_{z_2 \in \mc{V}_\mr{L}} \cdots \sum_{z_{k-1} \in \mc{V}_\mr{L}} \prx{\ms{F}_{<k}, Z_{<k} = z_{<k} \mid \ms{R}} \label{eq:lemma_total_prob}\\
		&= \prod_{i=1}^{k-1} \sum_{z \in \mc{V}_\mr{L}} \prx{\ms{F}_i, Z_i=z \mid \ms{R}}\label{eq:lemma_indep}\\
		&= \prod_{i=1}^{k-1} \sum_{z \in \mc{V}_\mr{L}} \prx{\ms{F}_i \mid Z_i=z, \ms{R}}\prx{Z_i=z \mid \ms{R}} \\
		&= \prod_{i=1}^{k-1} \sum_{z\in \mc{V}_\mr{L}} \left(1 - \alpha(z)\right) q(z)\\
		&= \left(\sum_{z\in \mc{V}_\mr{L}} \left(1 - \alpha(z)\right) q(z)\right)^{k-1}\\
		&= \left(1 - \sum_{z\in \mc{V}_\mr{L}} \alpha(z)q(z)\right)^{k-1},
		\label{eq:lemma_a3}
	\end{align}
	where~\eqref{eq:lemma_total_prob} comes from the law of total probability, and~\eqref{eq:lemma_indep} holds because each candidate and its verification result are independent from others.
\end{itemize}
By substituting~\eqref{eq:lemma_a1},~\eqref{eq:lemma_a2}, and~\eqref{eq:lemma_a3} into~\eqref{eq:lemma_accept_first}, we obtain 
\begin{align}
	\prx{Y=x, \Omega \mid \ms{R}} &= \sum_{k=1}^K \alpha(x) q(x) \left(1 - \sum_{z\in\mc{V}_\mr{L}}\alpha(z) q(z)\right)^{k-1}\\
	&= \alpha(x) q(x) \sum_{k=1}^K \left(1 - \sum_{z\in\mc{V}_\mr{L}}\alpha(z) q(z)\right)^{k-1}\\
	&= \alpha(x) q(x) \frac{1-\left(1 - \sum_{z\in\mc{V}_\mr{L}}\alpha(z) q(z)\right)^K}{\sum_{z\in\mc{V}_\mr{L}}\alpha(z) q(z)}\\
	&= \alpha(x) q(x) \frac{1-\left(1 - \Gamma \right)^K}{\Gamma},
	\label{eq:lemma_1st_cond}
\end{align}
where $\Gamma \triangleq \sum_{x'\in\mc{V}_\mr{L}} \alpha(x')q(x')$ is the probability that a replacement candidate is accepted. 
Note that we consider the nontrivial case $\Gamma >0$.
If $\Gamma = 0$, no replacement candidate is ever accepted, so $\Omega$ occurs with probability zero and SR-DV always falls back to HR, which samples exactly from $p_\mr{res}(x)$.
Therefore, it suffices to consider the case $\Gamma>0$ in the remainder of the proof.

\paragraph{2) Computing $\prx{Y=x, \Omega^c \mid  \ms{R}}$ in~\eqref{eq:lemma_total}}
Since $\Omega^c$ is the event that all $K$ candidates are rejected, we can rewrite $\prx{Y=x,\Omega^c \mid \ms{R}}$ as 
\begin{align}
	\prx{Y=x, \Omega^c \mid \ms{R}}
	&= \prx{Y=x, \ms{F}_{<K+1} \mid \ms{R}}\\
	&= \prx{Y=x \mid \ms{F}_{<K+1}, \ms{R}}\prx{\ms{F}_{<K+1} \mid  \ms{R}}.
\end{align}
By the design of the resampling procedure in SR-DV, when all $K$ replacement candidates are rejected, the algorithm falls back to HR.
Since HR samples the output token exactly from the residual distribution $p_\mr{res}(x)$, we have
\begin{align}
	\prx{Y=x \mid \ms{F}_{<K+1}, \ms{R}} = p_\mr{res}(x).
\end{align}
Moreover, $\prx{\ms{F}_{<K+1} \mid \ms{R}} = (1-\Gamma)^K$ follows directly from~\eqref{eq:lemma_a3}.
Therefore,
\begin{align}
	\prx{Y=x, \Omega^c \mid \ms{R}}
	= p_\mr{res}(x)\left(1 - \Gamma\right)^K.
	\label{eq:lemma_2nd_cond}
\end{align}

\paragraph{3) Combining $\prx{Y=x, \Omega \mid  \ms{R}}$ and $\prx{Y=x, \Omega^c \mid \ms{R}}$}
Using~\eqref{eq:lemma_1st_cond} and~\eqref{eq:lemma_2nd_cond}, we can obtain~\eqref{eq:lemma_total} as follows.
\begin{align}
	\prx{Y=x \mid \ms{R}} = \alpha(x) q(x) \frac{1-\left(1 - \Gamma \right)^K}{\Gamma} + p_\mr{res}(x)\left(1 - \Gamma\right)^K.
\end{align}

\paragraph{4) Deriving $\alpha(x)$} To ensure $\prx{Y=x \mid \ms{R}} = p_\mr{res}(x)$ for all $x \in \mc{V}_\mr{L}$, it is necessary and sufficient that 
\begin{align}
	\alpha(x) q(x) \frac{1-\left(1 - \Gamma \right)^K}{\Gamma}  = p_\mr{res}(x) \left(1-\left(1 - \Gamma \right)^K\right). 
	\label{eq:lemma_alpha_cond}
\end{align}

Since $\Gamma \in (0,1]$, we have $1-(1-\Gamma)^{K} > 0$. 
Moreover, $\alpha(x)$ only needs to be specified for tokens with $q(x)>0$, 
as a token with $q(x)=0$ is never sampled by the server. Thus,~\eqref{eq:lemma_alpha_cond} becomes
\begin{align}
	\alpha(x) = \Gamma\,\frac{p_{\mathrm{res}}(x)}{q(x)}.
	\label{eq:lemma_alpha_ratio}
\end{align}
Equation~\eqref{eq:lemma_alpha_ratio} characterizes a family of feasible acceptance functions parameterized by $\Gamma$. 
That is, for a given $\Gamma$, the acceptance function $\alpha(x)$ is determined accordingly.
This construction is self-consistent, since substituting~\eqref{eq:lemma_alpha_ratio} into $\Gamma=\sum_{x'\in \mc{V}_\mr{L}} \alpha(x')q(x')$ recovers $\Gamma$.

By the definition of $p_\mr{res}(x)$,~\eqref{eq:lemma_alpha_ratio} becomes
\begin{align}
	\alpha(x) 
	= \frac{\Gamma}{\theta} \frac{\max(q(x)-p^\mr{c}(x),0)}{q(x)},
	\label{eq:lemma_derived_alpha}
\end{align}
where $\theta \triangleq \sum_{x'\in\mc{V}_\mr{L}}\max(q(x')-p^\mr{c}(x'),0)$.

Next, since $\alpha(x)$ is an acceptance probability, it must satisfy
\begin{align}
	0 \leq \alpha(x) \leq 1, \quad \forall x \in \mc{V}_\mr{L}.
\end{align}
Given $\theta$ and $\Gamma$, $\alpha(x)$ is maximized for tokens satisfying $p^\mr{c}(x)=0$, for which~\eqref{eq:lemma_derived_alpha} reduces to $\frac{\Gamma}{\theta}$. This case includes LLM-only tokens, which are not represented in the device-side common-vocabulary distribution.
Thus, the feasibility condition $\alpha(x) \leq 1$ implies
\begin{align}
	\Gamma \leq \theta.
\end{align}
Since $\Gamma$ is the probability that a replacement candidate is accepted, a larger $\Gamma$ leads to a smaller fallback probability $\prx{\ms{F}_{<K+1} \mid \ms{R}}=(1-\Gamma)^K$, thereby reducing the frequency of invoking HR and the corresponding communication overhead.
Hence, minimizing the communication overhead is equivalent to taking the largest $\Gamma$ under the feasibility constraint $\alpha(x) \leq 1$, i.e., 
\begin{align}
	\Gamma^* = \theta.
\end{align}
Substituting $\Gamma^*$ into~\eqref{eq:lemma_derived_alpha} yields
\begin{align}
	\alpha(x) = \frac{\max(q(x)-p^\mr{c}(x), 0)}{q(x)},
\end{align}
which completes the proof.

\subsection{Proof of Theorem~2}\label{appendix:thm2}
For all $x \in \mc{V}_\mr{L}$, we have
\begin{align}
	\prx{Y=x} = \prx{Y=x,\ms{A}} + \prx{Y=x,\ms{R}}
	\label{eq:srdv_total}.
\end{align}

\paragraph{1) Computing $\prx{Y=x,\ms{A}}$ in~\eqref{eq:srdv_total}}
We can obtain $\prx{Y=x,\ms{A}}$ as follows.
\begin{align}
	\prx{Y=x, \ms{A}} &= \prx{X=x, \ms{A}}\\
	&= \prx{\ms{A} \mid X=x} \prx{X=x} \\
	&= \min\left(1,\frac{q(x)}{p^\mr{c}(x)}\right) p^\mr{c}(x)\\
	&= \min(p^\mr{c}(x), q(x)).
	\label{eq:srdv_term1}
\end{align}

\paragraph{2) Computing $\prx{Y=x,\ms{R}}$ in~\eqref{eq:srdv_total}}
By the chain rule, we have
\begin{align}
	\prx{Y=x,\ms{R}} = \prx{Y=x \mid \ms{R}} \prx{\ms{R}}.
\end{align}
By \textbf{Lemma~\ref{lemma}}, under $\alpha(x)=\frac{\max(q(x)-p^\mr{c}(x),0)}{q(x)}$, the conditional output distribution under rejection satisfies $\prx{Y=x \mid \ms{R}} = p_\mr{res}(x)$. 
Next, we can obtain $\prx{\ms{R}}$ as follows.
\begin{align}
	\prx{\ms{R}} &= \sum_{x\in\mc{V}_\mr{L}} \prx{X=x, \ms{R}}\\
	&= \sum_{x\in\mc{V}_\mr{L}} \prx{\ms{R} \mid X=x} \prx{X=x}\\
	&= \sum_{x \in \mc{V}_\mr{L}} \left(1 - \min\left(1,\frac{q(x)}{p^\mr{c}(x)}\right)\right) p^\mr{c}(x)\\
	&= \sum_{x \in \mc{V}_\mr{L}} \left(p^\mr{c}(x) - \min(p^\mr{c}(x),q(x))\right) \\
	&= \sum_{x \in \mc{V}_\mr{L}} \left(q(x) - \min(p^\mr{c}(x),q(x))\right)  \\
	&= \sum_{x\in \mc{V}_\mr{L}} \max(q(x)-p^\mr{c}(x),0).
\end{align} 
Therefore, 
\begin{align}
	\prx{Y=x,\ms{R}} &= p_\mr{res}(x) \sum_{x'\in \mc{V}_\mr{L}} \max(q(x')-p^\mr{c}(x'),0)\\
	&= \max(q(x)-p^\mr{c}(x),0).
	\label{eq:srdv_term2}
\end{align}

\paragraph{3) Combining $\prx{Y=x,\ms{A}}$ and $\prx{Y=x,\ms{R}}$}
Substituting~\eqref{eq:srdv_term1} and~\eqref{eq:srdv_term2} into~\eqref{eq:srdv_total}, we can obtain
\begin{align}
	\prx{Y=x} &= \min(p^\mr{c}(x), q(x)) + \max(q(x)-p^\mr{c}(x),0)\\
	&= q(x),
\end{align}
which completes the proof.

\section{Implementation details}\label{appendix:implementation}
\paragraph{Hardware and detailed settings} 
We conduct all experiments on a single NVIDIA A100 40GB GPU.
To emulate the distributed CoSD setting, the on-device computing time is scaled according to the measured generation-speed ratio of the on-device SLM on an NVIDIA TITAN RTX relative to that on the NVIDIA A100 GPU.
Following~\cite{timor25, ramakrishnan25}, we use the Hugging Face models double7/vicuna-68m as the on-device SLM and Qwen/Qwen2-7B and meta-llama/Llama-3.1-8B as the server LLMs.
Similar to~\cite{ramakrishnan25}, we construct the common vocabulary $\mc{V}_\mr{C}$ using tokens that admit a one-to-one direct mapping between the SLM and the LLM.
The vocabulary overlap between the on-device SLM and the server LLMs is reported in Table~\ref{tab:common_ratio}.
\begin{table}[!h]
	\caption{Vocabulary overlap between vicuna-68m ($|\mc{V}_\mr{S}|=32,000$) and the server LLMs}
	\begin{center}
		\begin{tabular}{lccc}
			\toprule
			Server LLM type & $|\mc{V}_\mr{L}|$ & $|\mc{V}_\mr{C}|$ & $\frac{|\mc{V}_\mr{C}|} {|\mc{V}_\mr{L}|}$\\
			\midrule
			Qwen2-7B & 151,646 & 22,269 & 0.1468\\
			Llama-3.1-8B & 128,256 & 22,413 & 0.1748\\
			\bottomrule
		\end{tabular}
	\end{center}
	\label{tab:common_ratio}
\end{table}
\paragraph{Baselines} 
We compare the proposed methods with Naive, UL, GR, TR, U-HLM, and server LLM. 
Among the baselines, U-HLM requires an uncertainty measure to determine which candidate tokens are verified. 
Specifically, the uncertainty $u(x)$ of a candidate token $x \in \mc{V}_\mr{C}$ is measured as 
\begin{align}
	u(x) = \frac{\sum_{e=1}^E \mathbb{I}(x_{(e)} \neq x)}{E},
\end{align}
where $E$ is the number of softmax temperatures sampled from $[0,\eta_\mr{max}]$,  and $x_{(e)}$ represents a token sampled from the perturbed SLM distribution at the $e$-th temperature. 
Here, $\mathbb{I}(\cdot)$ is an indicator function.
Only the tokens with $u(x) > u_\mr{th}$ are verified. 
We set $E=20$, $\eta_\mr{max}=2.0$, and $u_\mr{th}=0.5$ in our experiments.

The detailed communication load per verification round of all methods except server LLM is summarized in Table~\ref{tab:comm_load}. Note that, since all methods transmit only the bonus token when all candidate tokens are accepted,  we report the downlink communication load only for the rejection case.
\begin{table}[!h]
	\centering
	\caption{Communication load per verification round}
	\label{tab:comm_load}
	{\footnotesize
	\begin{tabular}{lcc}
		\toprule
		Method & Uplink (device-to-server) & Downlink (rejection, server-to-device) \\
		\midrule
		X-CoSD
		& candidate tokens and probabilities
		& LLM distribution over $\mc{V}_\mr{C}$ and $\theta_\mr{o}$ \\
		
		\multirow{2}{*}{X-CoSD-E}
		& \multirow{2}{*}{candidate tokens and probabilities}
		& $K$ replacement candidates and probabilities \\
		& & fallback: LLM distribution over $\mc{V}_\mr{C}$ and $\theta_\mr{o}$ \\
		\midrule
		
		Naive
		& candidate tokens and probabilities
		& full LLM distribution over $\mc{V}_\mr{L}$ \\
		
		UL
		& candidate tokens and SLM distributions over $\mc{V}_\mr{C}$
		& resampled token \\
		
		GR / TR
		& candidate tokens and probabilities
		& resampled token \\
		
		U-HLM
		& uncertain candidate token and SLM distribution over $\mc{V}_\mr{C}$
		& resampled token \\
		\bottomrule
	\end{tabular}}
\end{table}

\section{Additional experiment results}
\subsection{Communication load per token on additional datasets}\label{appendix:load}
We report communication load per generated token on CNN/DailyMail, GSM8K, and MMLU datasets in Table~\ref{tab:load_per_token_appendix}. 
Consistent with the results in Table~\ref{tab:load_per_token}, X-CoSD and X-CoSD-E achieve substantially lower downlink communication load than Naive across all datasets and server LLMs.
In contrast, UL and U-HLM still incur large uplink communication overhead.
These results further support the practicality of X-CoSD and X-CoSD-E for collaborative inference under limited wireless communication resources and heterogeneous SLM--LLM vocabularies.
\begin{table}[!h]
	\centering
	\caption{Communication load per generated token for $N=2$ (bits)}
	{\footnotesize
		\begin{tabular}{llcccccc}
			\toprule
			\multirow{2}{*}{LLM type} & \multirow{2}{*}{Method} & \multicolumn{2}{c}{CNN/DailyMail} & \multicolumn{2}{c}{GSM8K} & \multicolumn{2}{c}{MMLU}\\
			\cmidrule(lr){3-4}
			\cmidrule(lr){5-6}
			\cmidrule(lr){7-8}
			& & Uplink & Downlink & Uplink & Downlink & Uplink & Downlink\\
			\midrule
			\multirow{5}{*}{Qwen2-7B} & \textbf{X-CoSD (prop.)} & \textbf{122.2} & \textbf{0.2 M} & \textbf{131.4} & \textbf{0.2 M} & \textbf{66.0} & \textbf{0.2 M}\\
			& \textbf{X-CoSD-E (prop.)} & \textbf{117.9} & \textbf{335.1} & \textbf{125.0} & \textbf{383.0} & \textbf{61.0} & \textbf{341.5}\\
			& Naive & 117.5 & 1.0 M & 125.0 & 1.2 M & 61.3 & 1.1 M\\
			& UL & 0.2 M & 10.4 & 0.2 M & 11.3 & 0.2M & 10.5\\
			& U-HLM ($u_\mr{th}=0.5$)& 0.1 M & 4.8 & 0.1 M & 4.7 & 0.1M & 6.2\\
			\midrule
			\multirow{5}{*}{Llama-3.1-8B} & \textbf{X-CoSD (prop.)} & \textbf{114.2} & \textbf{0.1 M} & \textbf{127.9} & \textbf{0.2 M} & \textbf{72.9} & \textbf{0.1 M}\\
			& \textbf{X-CoSD-E (prop.)} & \textbf{120.1} & \textbf{251.9} & \textbf{123.4} & \textbf{335.2} & \textbf{68.5} & \textbf{294.5}\\
			& Naive & 111.7 & 0.6M  & 123.4 & 0.9 M & 68.8 & 0.8 M\\
			& UL & 0.2 M & 8.6 & 0.2 M & 9.9 & 0.2 M & 9.1\\
			& U-HLM ($u_\mr{th}=0.5$)& 0.1 M & 4.0 & 0.1 M &  4.2 & 0.1 M &  4.9\\
			\bottomrule
		\end{tabular}
	}
	\label{tab:load_per_token_appendix}
\end{table}

\subsection{Token throughput on additional datasets}\label{appendix:tput}
\begin{figure}[h]
	\centering
	\begin{subfigure}[h]{\columnwidth}
		\centering
		\includegraphics[width=0.32\linewidth]{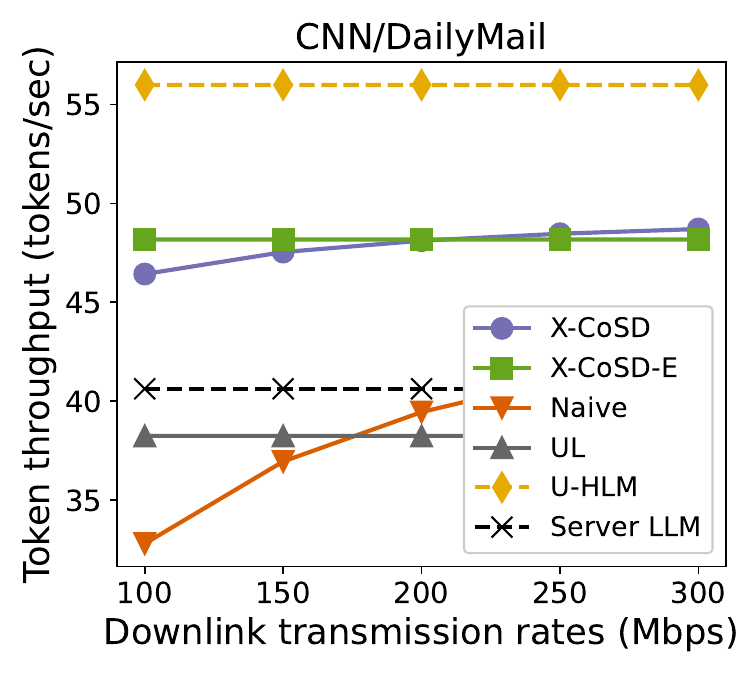}
		\includegraphics[width=0.32\linewidth]{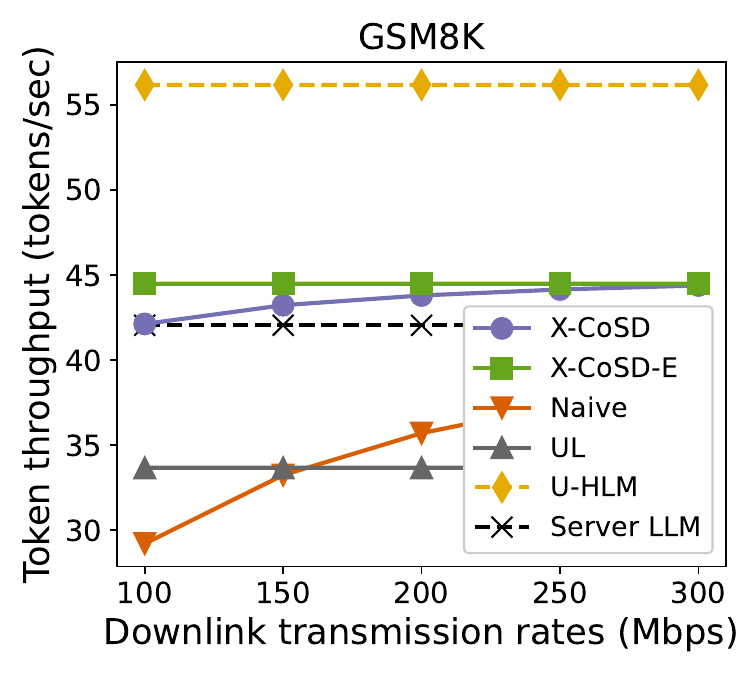}
		\includegraphics[width=0.32\linewidth]{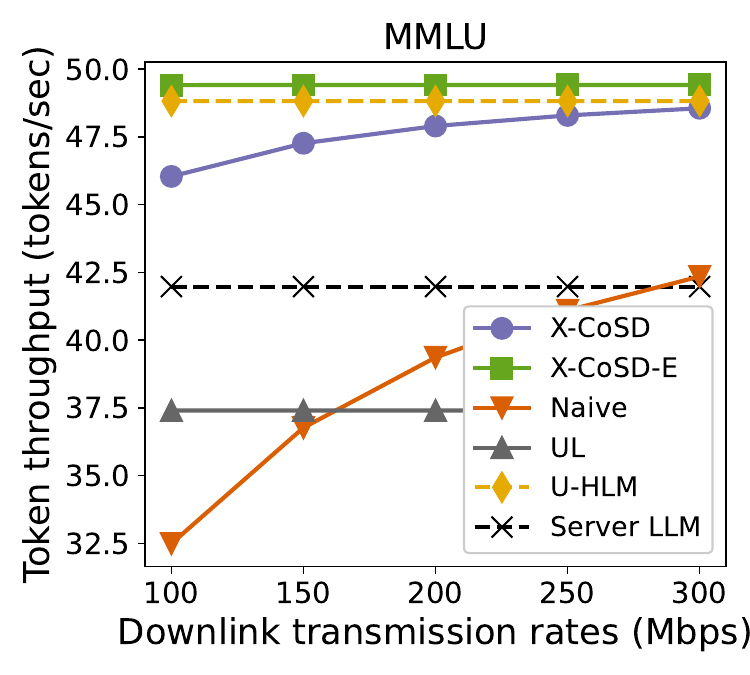}
		\caption{Qwen2-7B}
	\end{subfigure}
	\centering
	\begin{subfigure}[h]{\columnwidth}
		\centering
		\includegraphics[width=0.32\linewidth]{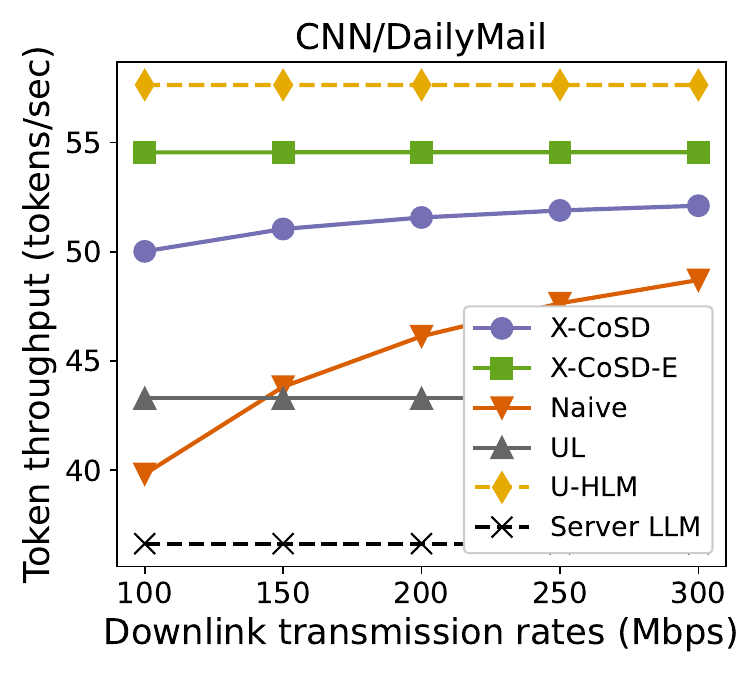}
		\includegraphics[width=0.32\linewidth]{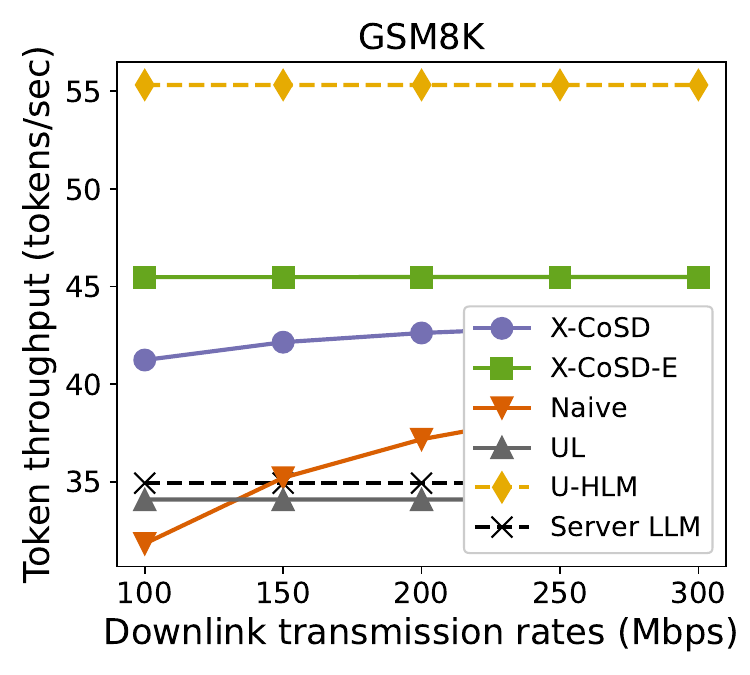}
		\includegraphics[width=0.32\linewidth]{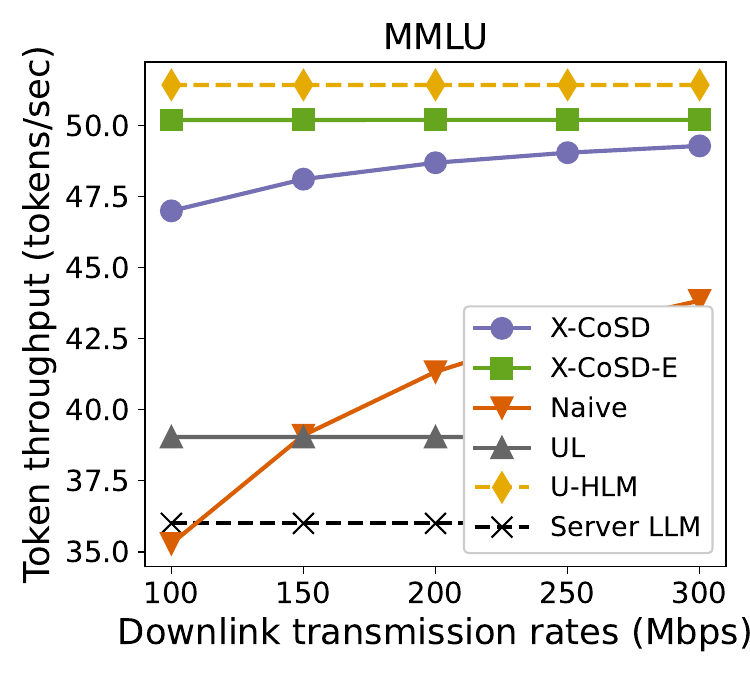}
		\caption{Llama-3.1-8B}
	\end{subfigure}
	\caption{Token throughput versus the downlink transmission rates.}
	\label{fig:sim_tput_appendix}
\end{figure}

Figure~\ref{fig:sim_tput_appendix} presents token throughput for the  CNN/DailyMail, GSM8K, and MMLU datasets. 
The results show the same trend as in Figure~\ref{fig:sim_tput_wmt_xsum}.
In particular, X-CoSD and X-CoSD-E achieve higher token throughput than Server LLM, Naive, and UL by reducing downlink communication overhead.
Moreover, X-CoSD-E consistently shows high token throughput regardless of downlink transmission rates thanks to SR-DV.
Note that, although U-HLM attains high token throughput, it does not guarantee lossless decoding.

\subsection{Per-token latency on additional datasets}\label{appendix:cdf}
\begin{figure}[!h]
	\centering
	\begin{subfigure}[h]{0.95\columnwidth}
		\centering
		\includegraphics[width=0.49\linewidth]{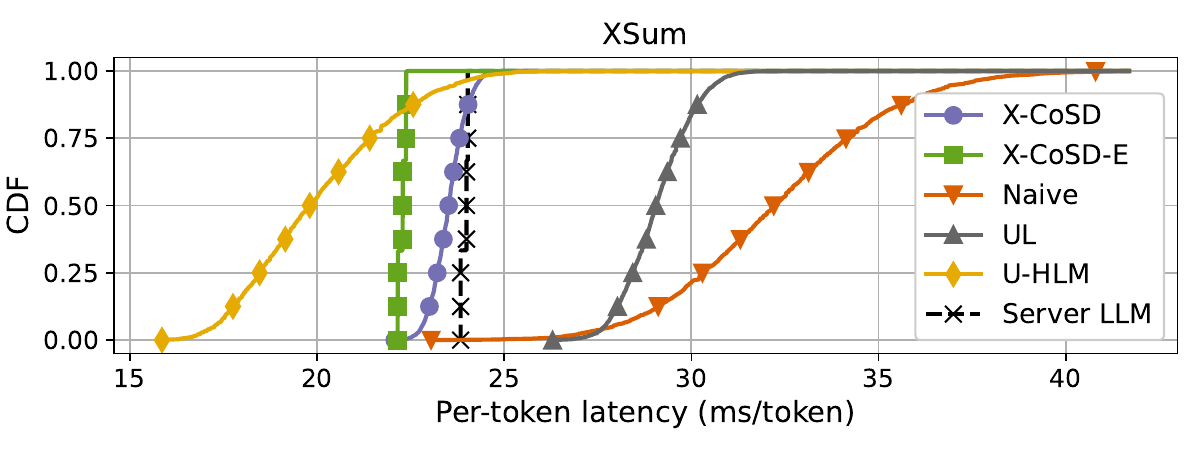}
		\includegraphics[width=0.49\linewidth]{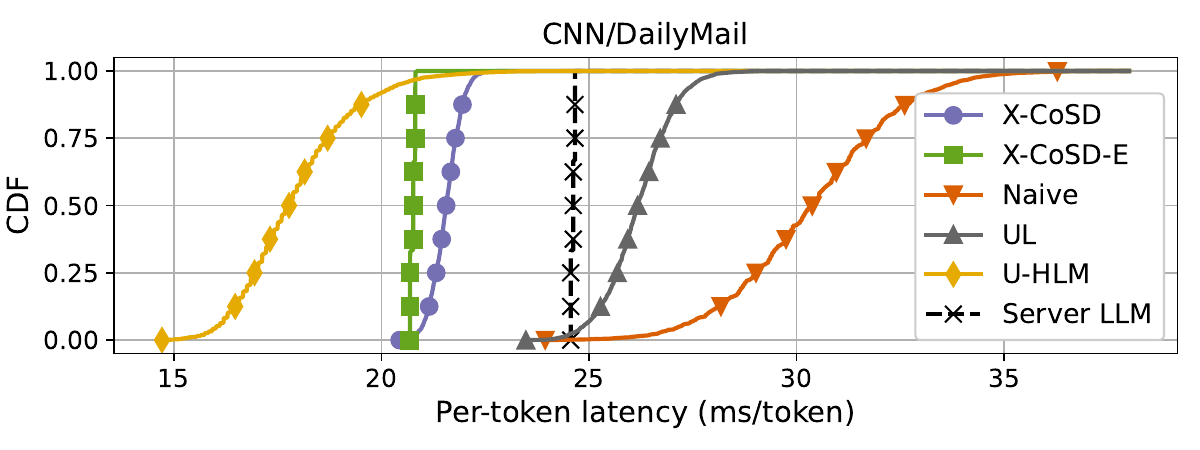}
		\includegraphics[width=0.49\linewidth]{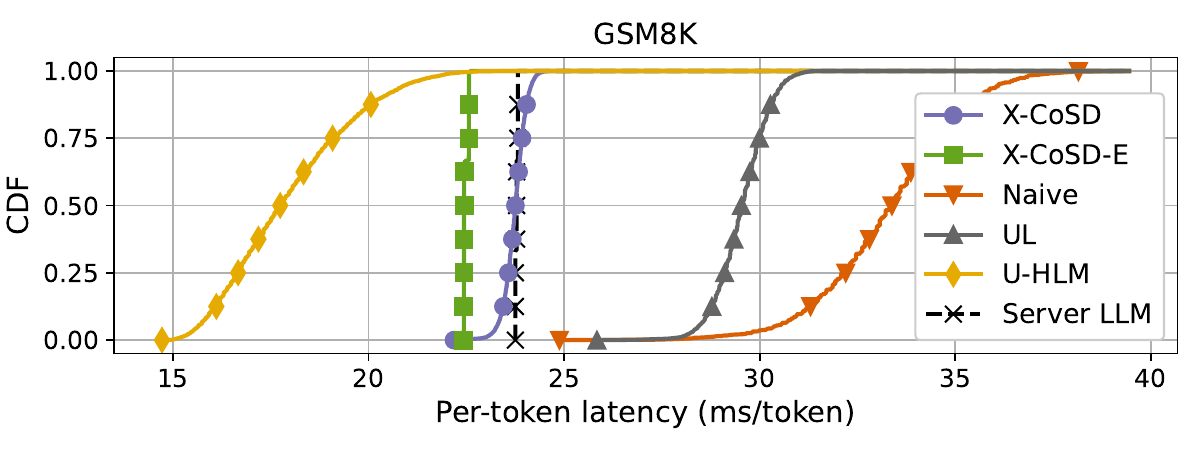}
		\includegraphics[width=0.49\linewidth]{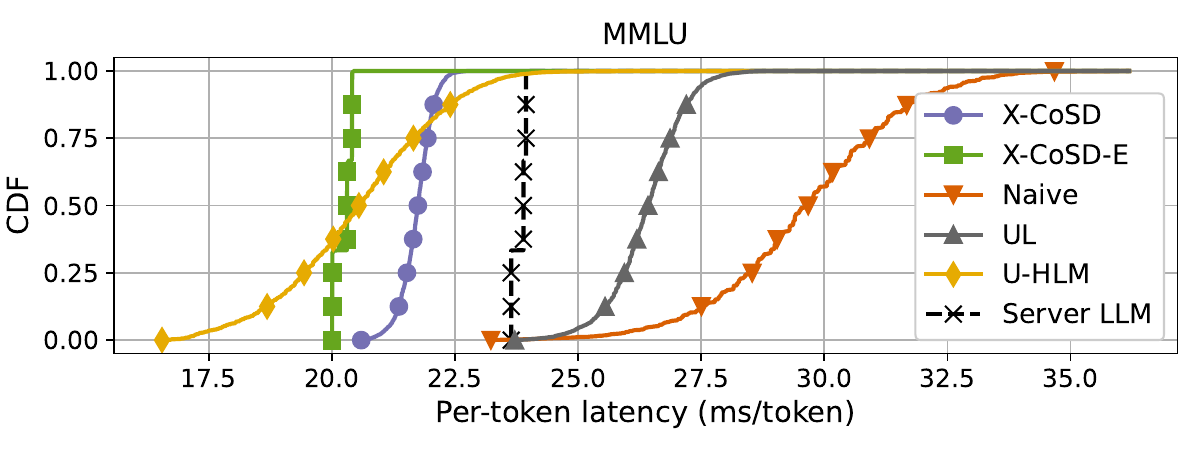}
		\caption{Qwen2-7B}
		\label{fig:cdf_qwen_appendix}
	\end{subfigure}
	\centering
	\begin{subfigure}[h]{0.95\columnwidth}
		\centering
		\includegraphics[width=0.49\linewidth]{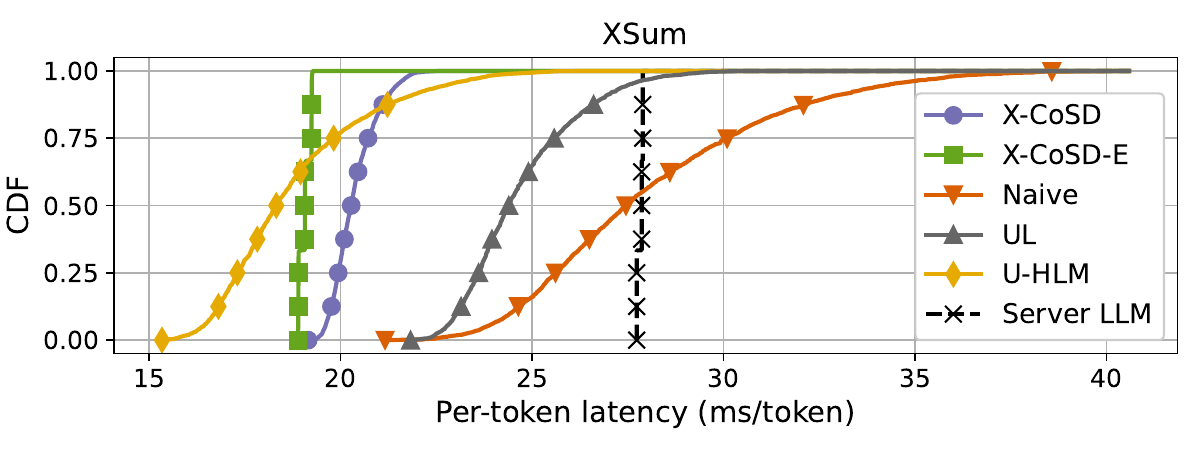}
		\includegraphics[width=0.49\linewidth]{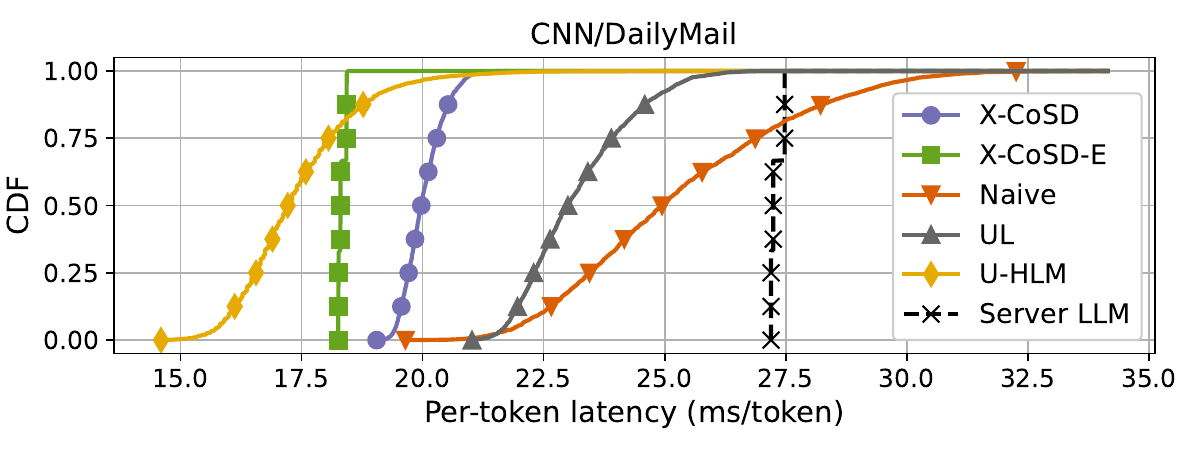}
		\includegraphics[width=0.49\linewidth]{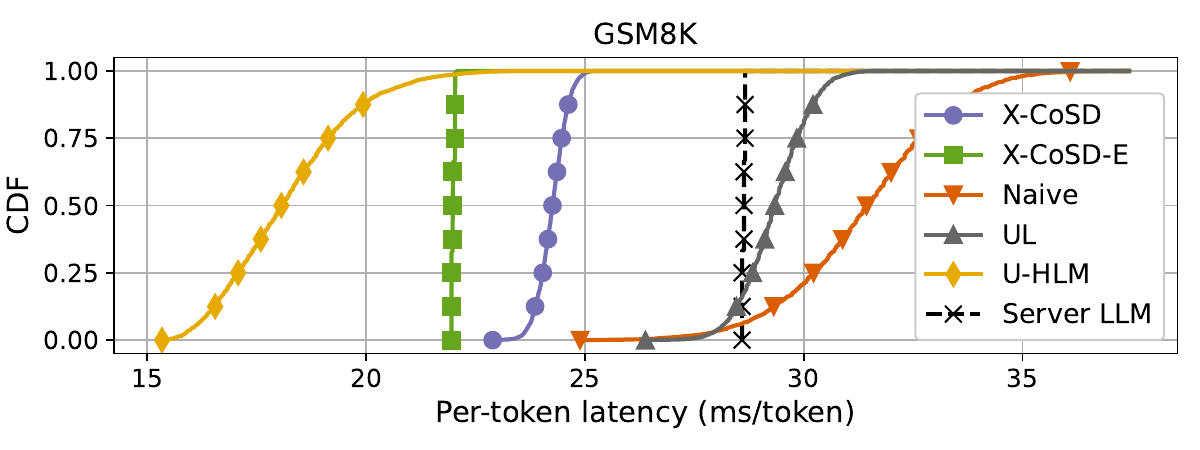}
		\includegraphics[width=0.49\linewidth]{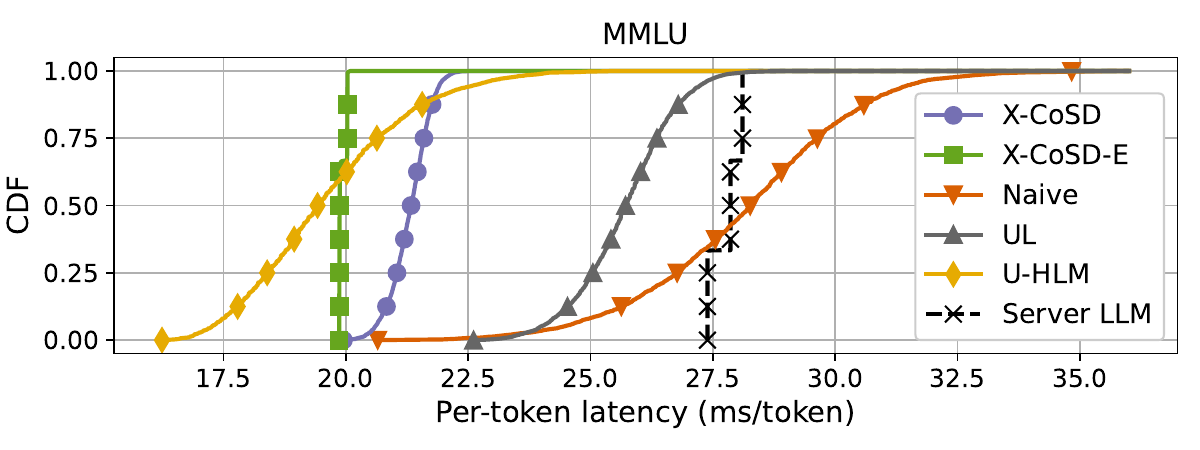}
		\caption{Llama-3.1-8B}
		\label{fig:cdf_llama_appendix}
	\end{subfigure}
	\caption{CDF of per-token latency.}
	\label{fig:sim_cdf_appendix}
\end{figure}

Figure~\ref{fig:sim_cdf_appendix} shows per-token latency for the XSum, CNN/DailyMail, GSM8K, and MMLU datasets. 
The results show the same trend as in Figure~\ref{fig:sim_cdf_wmt}.
The proposed X-CoSD and X-CoSD-E show lower per-token latency than other baselines, except for the lossy baseline U-HLM.
Although U-HLM attains lower per-token latency, it does not guarantee lossless decoding.
Moreover, by significantly reducing communication overhead, X-CoSD-E consistently shows lower per-token latency than X-CoSD.

\section{Limitations}\label{appendix:limitation}
A key limitation of this work is that, under the TLI~\cite{timor25}-based formulation, the on-device SLM can generate candidate tokens only from the common vocabulary shared with the server LLM.
As a result, the performance may depend on the size of the common vocabulary, since both the computational efficiency of SD and the communication load between the user device and the edge server can vary accordingly.
Finally, our experiments are limited to two server LLMs, Qwen2-7B and Llama-3.1-8B.
Nevertheless, we evaluated the proposed frameworks across diverse tasks under challenging low-overlap settings, where the common vocabulary size is less than 20\% of the full LLM vocabulary.
Developing CoSD methods that mitigate dependence on the common vocabulary while preserving lossless decoding remains an important direction for future work.

\end{document}